\documentclass[10pt,journal,compsoc]{IEEEtran}

\ifCLASSOPTIONcompsoc
  \usepackage[nocompress]{cite}
\else
  \usepackage{cite}
\fi

\ifCLASSINFOpdf
\else
\fi

\usepackage{bm}
\usepackage{amsmath}
\usepackage{amssymb}
\usepackage{multirow}
\usepackage{subfigure}
\usepackage{array}
\usepackage{amsthm}
\usepackage{amsfonts}
\usepackage{graphicx}
\usepackage{rotating}
\usepackage{pifont}
\usepackage[linesnumbered,ruled]{algorithm2e}

\newcommand{\citet}{\cite}
\newcommand{\figref}[1]{Fig. \ref{#1}}
\newcommand{\tabref}[1]{Table \ref{#1}}
\newcommand{\secref}[1]{Section \ref{#1}}
\newcommand{\appendixref}[1]{the Appendix}
\newcommand{\equref}[1]{Equation (\ref{#1})}
\newcommand{\algoref}[1]{Algorithm \ref{#1}}

\begin{document}
%
\title{Label-Enhanced Graph Neural Network for Semi-supervised Node Classification}
%
%
%

\author{Le~Yu,
        Leilei~Sun,
        Bowen~Du,
        Tongyu~Zhu,
        Weifeng~Lv
\IEEEcompsocitemizethanks{\IEEEcompsocthanksitem L. Yu, L. Sun, B. Du, T. Zhu and W. Lv are with the State Key Laboratory of Software Development Environment, Beihang University, Beijing, 100191, China.\protect\\
E-mail: \{yule,leileisun,dubowen,zhutongyu,lwf\}@buaa.edu.cn
}

\thanks{(Corresponding author: Tongyu Zhu.)}}

%
%

\markboth{IEEE TRANSACTIONS ON KNOWLEDGE AND DATA ENGINEERING,~Vol.~XX, No.~X, XX~XXXX}%
{Yu \MakeLowercase{\textit{et al.}}: Label-Enhanced Graph Neural Network for Semi-supervised Node Classification}
%

\IEEEtitleabstractindextext{%
\begin{abstract}
Graph Neural Networks (GNNs) have been widely applied in the semi-supervised node classification task, where a key point lies in how to sufficiently leverage the limited but valuable label information. Most of the classical GNNs solely use the known labels for computing the classification loss at the output. In recent years, several methods have been designed to additionally utilize the labels at the input. One part of the methods augment the node features via concatenating or adding them with the one-hot encodings of labels, while other methods optimize the graph structure by assuming neighboring nodes tend to have the same label. To bring into full play the rich information of labels, in this paper, we present a label-enhanced learning framework for GNNs, which first models each label as a virtual center for intra-class nodes and then jointly learns the representations of both nodes and labels. Our approach could not only smooth the representations of nodes belonging to the same class, but also explicitly encode the label semantics into the learning process of GNNs. Moreover, a training node selection technique is provided to eliminate the potential label leakage issue and guarantee the model generalization ability. Finally, an adaptive self-training strategy is proposed to iteratively enlarge the training set with more reliable pseudo labels and distinguish the importance of each pseudo-labeled node during the model training process. Experimental results on both real-world and synthetic datasets demonstrate our approach can not only consistently outperform the state-of-the-arts, but also effectively smooth the representations of intra-class nodes.
\end{abstract}

}

\maketitle

\IEEEdisplaynontitleabstractindextext

%
\IEEEpeerreviewmaketitle

\IEEEraisesectionheading{\section{Introduction}
\label{section-1}}

\IEEEPARstart{G}{raphs} are ubiquitous in the real world, which represent the objects and their relationships as nodes and edges, respectively \cite{DBLP:journals/tkde/ZhangCZ22}. One fundamental learning task on graphs is the semi-supervised node classification task, which plays an essential role in various applications such as predicting the areas of publications in academic networks \cite{DBLP:conf/iclr/KipfW17}, inferring the categories of products in co-purchasing networks \cite{DBLP:conf/kdd/ChiangLSLBH19}, and identifying the functions of proteins in biology graphs \cite{DBLP:conf/nips/HamiltonYL17}. Semi-supervised node classification aims to predict the labels of unlabeled nodes given a partially labeled graph, where a key point for achieving satisfactory performance is how to comprehensively utilize the limited but valuable label information.

\begin{figure}[!ht]
    \centering
    \includegraphics[width=\columnwidth]{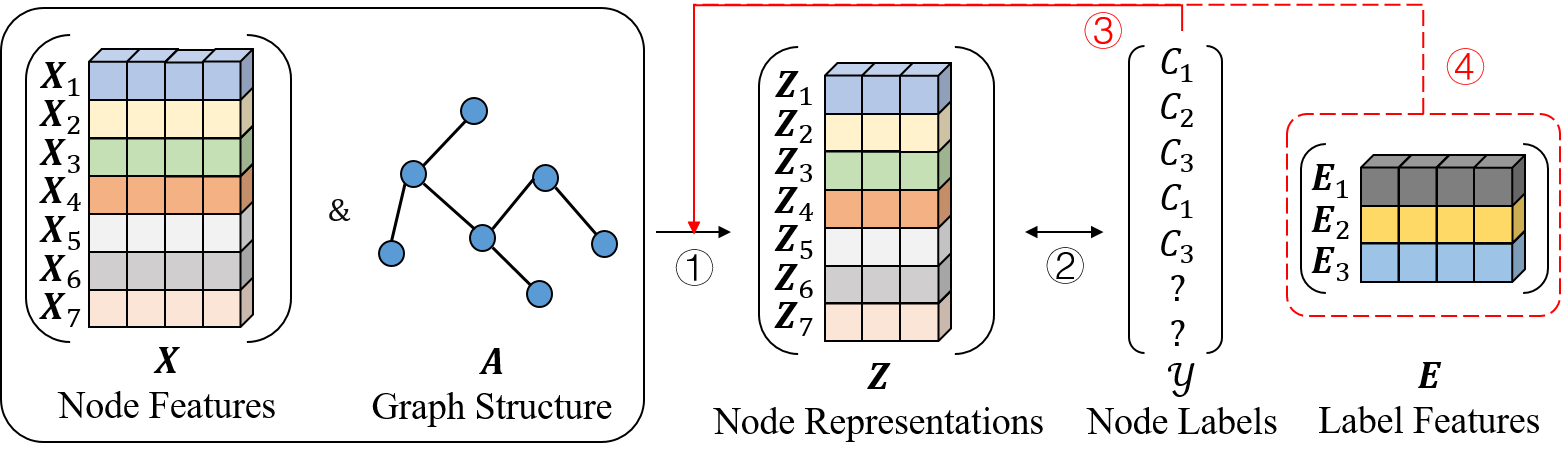}
    \caption{Illustration of the proposed semi-supervised learning framework that enhances GNNs with rich label information. We incorporate the label information into the learning process of GNNs via \ding{174}, and we also encode the label semantic features through \ding{175}.}
\label{fig:label_usage}
\end{figure}

Recent years have witnessed the success of Graph Neural Networks (GNNs) on the semi-supervised node classification task because they could learn node representations with the consideration of both node features and graph structure simultaneously \cite{DBLP:journals/tnn/WuPCLZY21}. For the usage of labels, most of the popular GNNs (e.g., GCN \cite{DBLP:conf/iclr/KipfW17}, GraphSAGE \cite{DBLP:conf/nips/HamiltonYL17} and GAT \cite{DBLP:conf/iclr/VelickovicCCRLB18}) learn the mapping function between node representations and labels, where labels are only leveraged for computing the classification loss \textit{at the output} in \ding{173} in \figref{fig:label_usage}. It can be found that labels cannot be explicitly accessed by the node representation learning process in \ding{172}, which hinders GNNs from comprehensively considering the rich information of labels. 

To further improve the performance of GNNs, some recent efforts attempted to incorporate the label information into the node representation learning process in GNNs via \ding{174}. These methods either augmented the node features or optimized the graph structure based on labels \textit{at the input}. In particular, \citet{DBLP:journals/corr/abs-2103-13355} and \citet{DBLP:conf/ijcai/ShiHFZWS21} augmented the node features with the one-hot encodings of labels through the concatenation and addition operations, respectively. \citet{chen2019highwaygraph,DBLP:conf/ijcnn/YangYDCC21} leveraged labels to modify the graph structure via adding intra-class edges and removing inter-class edges among nodes. \citet{DBLP:conf/ijcai/0002KCJ0G19,wang2021combining} optimized the graph structure by Label Propagation Algorithm (LPA) \cite{DBLP:conf/nips/ZhouBLWS03,DBLP:conf/icml/ZhuGL03} with the assumption that nearby nodes tend to share the same label. Although these methods utilize labels to improve the learning of node representations, the rich information of labels (explained in the next paragraph) is still not fully exploited.

In fact, labels can carry valuable information which is beneficial for classifying nodes. Firstly, each label can be treated as a virtual center for nodes belonging to the label and reflects the intra-class node proximity. For example, in academic networks, papers in the same area are more relevant than those in different areas. In business networks, products with the same category tend to reflect similar characteristics. Secondly, labels are associated with rich semantics and some labels can be semantically close to each other. For instance, the Artificial Intelligence and Machine Learning areas are more interrelated than the Artificial Intelligence and Chemistry areas. The relationships of computers and mouses are closer than those of computers and digital cameras. Therefore, when classifying the areas of papers or the categories of products, it is essential to explore the above abundant information of labels, which motivates us to design a new framework to improve the performance of GNNs in semi-supervised node classification by fully considering the label information.

In this paper, we propose a \textbf{L}abel-\textbf{E}hanced \textbf{G}raph \textbf{N}eural \textbf{N}etwork (LEGNN) to comprehensively integrate the rich information of labels into GNNs for facilitating semi-supervised node classification.
Concretely, we first construct a heterogeneous graph by creating a new type of nodes for labels with the semantic features and establishing their connections with intra-class nodes to make each label serve as the center of the corresponding nodes. Then, we design a general heterogeneous message passing mechanism to jointly learn the representations of both nodes and labels, which can effectively smooth intra-class node representations and explicitly encode label semantics. Moreover, we present a training node selection technique to cope with the potential label leakage issue and guarantee the model generalization ability. Finally, an adaptive self-training strategy is designed to iteratively enlarge the training set with more reliable pseudo labels and distinguish the importance of each pseudo-labeled node based on the training and evaluating confidence. We conduct extensive experiments on both real-world and synthetic datasets to validate the effectiveness of our method. Experimental results show that our approach can consistently outperform the existing methods and effectively enhance the smoothness of the representations of intra-class nodes.
The contributions of this paper are summarized as follows:

\begin{itemize}
    \item
    \textit{A label-enhanced learning framework} is proposed, which can fully utilize the rich information carried by labels to improve the performance of GNNs in semi-supervised node classification. This framework exploits the role of labels in the learning process of GNNs, which is not investigated by previous studies.
    
    \item
    \textit{A heterogeneous message passing mechanism} is designed to realize the co-learning of both nodes and labels, which can explicitly encode label semantics and effectively smooth the intra-class node representations. This mechanism is general and applicable to any message passing GNN.

    \item
    \textit{An adaptive self-training strategy} is designed to provide more reliable pseudo labels and discriminate the importance of each pseudo-labeled node according to the training and evaluating confidence.
\end{itemize}

\section{Preliminaries}
\label{section-2}

A graph can be represented as $\mathcal{G}=\left(\mathcal{V},\mathcal{E}\right)$, where $\mathcal{V} = \left\{v_1, v_2, \cdots,v_M \right\}$ is the node set and $\mathcal{E}$ is the edge set. Nodes are associated with a feature matrix $\bm{X} \in \mathbb{R}^{M \times F}$, where $F$ is the number of node features.
Let $\mathcal{L}$ and $\mathcal{U}$ be the set of labeled nodes and unlabeled nodes, where $\mathcal{L} \ \cap \ \mathcal{U} = \varnothing$, $\mathcal{L} \ \cup \ \mathcal{U} = \mathcal{V}$.
The node label matrix $\bm{Y} \in \mathbb{R}^{M \times C}$ consists of one-hot encoding vectors for labeled nodes and zero vectors for unlabeled nodes, where $C$ is the number of label classes. 
Specifically, each labeled node $v_i \in \mathcal{L}$ has a one-hot vector $\bm{Y}_i \in \left\{0,1\right\}^C$, where the entry of 1 indicates the label class of $v_i$. For each unlabeled node $v_i \in \mathcal{U}$, $\bm{Y}_i \in \left\{0\right\}^C$ is a all-zero vector. 
Let $\bm{A} \in \mathbb{R}^{M \times M}$ be the adjacency matrix. $\bm{A}$ is a binary matrix, i.e., $A_{i,j}=1$ if an edge exists between node $v_i$ and node $v_j$, and 0 otherwise.

Given a graph $\mathcal{G}=\left(\mathcal{V},\mathcal{E}\right)$, a node feature matrix $\bm{X}$, a set of labeled nodes $\mathcal{L} \subseteq \mathcal{V}$, a node label matrix $\bm{Y}$ where $\bm{Y}_i \in \left\{0,1\right\}^C$ for each node $v_i \in \mathcal{L}$ and each node belongs to exactly one label class, \textbf{semi-supervised node classification} aims to predict the labels of unlabeled nodes in $\mathcal{U} = \mathcal{V} \ \backslash \ \mathcal{L}$. Compared with the supervised setting where only the labeled node data could be used, semi-supervised node classification allows the models to use both labeled and unlabeled node data to classify the unlabeled nodes.

Existing solutions for semi-supervised node classification primarily relies on two steps: 1) leverage the graph information (i.e., node features and graph structure) to obtain node representations $\bm{Z} \in \mathbb{R}^{M \times D}$ with $D$ as the hidden dimension; 2) learn the mapping function by taking $\bm{Z}$ as the input and provide the predicted probability $\hat{\bm{Y}} \in \mathbb{R}^{M \times C}$ over all the label classes. The following cross-entropy loss is widely adopted as the objective function for each node $v_i$,
\begin{equation}\label{equ:cross_entropy_loss}
L(\bm{Y}_i, \hat{\bm{Y}}_i)=- \sum_{c=1}^C Y_{ic}log(\hat{Y}_{ic}),
\end{equation}
where $Y_{ic}$ and $\hat{Y}_{ic}$ are the ground truth and predicted probability of label class $c$ for node $v_i$.

\section{Related Work}
\label{section-3}

This section reviews the existing related literature and also points out their differences with our work.

\subsection{Semi-supervised Node Classification with Graph Neural Networks}
Semi-supervised node classification is one of the most important tasks in graph learning. In recent years, GNNs have achieved superior performance on semi-supervised node classification \cite{DBLP:journals/corr/abs-1812-08434,DBLP:journals/tnn/WuPCLZY21,DBLP:conf/iclr/KipfW17,DBLP:conf/nips/HamiltonYL17,DBLP:conf/iclr/VelickovicCCRLB18}.
Compared with the traditional graph embedding methods that mainly focused on the graph structure (e.g., DeepWalk \cite{DBLP:conf/kdd/PerozziAS14}, LINE \cite{DBLP:conf/www/TangQWZYM15} and node2vec \cite{DBLP:conf/kdd/GroverL16}), GNNs can consider both node features and graph structure simultaneously. 

Formally, GNNs first perform message passing among nodes and their neighbors according to the graph structure, and then compute node representations by aggregating the received information. The calculation of a GNN with $K$ layers can be represented by
\begin{equation}\label{equ:input_gnn}
\bm{H}^0 = \bm{X},
\end{equation}
\begin{equation}\label{equ:convolution_gnn}
\bm{H}^{k+1}  = \sigma\left(\overline{\bm{A}}\bm{H}^k\bm{W}^k\right),
\end{equation}
\begin{equation}\label{equ:output_gnn}
\bm{Z}  = \bm{H}^K,
\end{equation}
where $\bm{W}^k$ is the feature transformation matrix at layer $k$, $\sigma(\cdot)$ denotes the activation function. $\overline{\bm{A}}$ denotes the GNN-specific adjacency matrix, e.g., the normalized adjacency matrix in GCN \cite{DBLP:conf/iclr/KipfW17}, the sampled adjacency matrix in GraphSAGE \cite{DBLP:conf/nips/HamiltonYL17} and the attention-based adjacency matrix in GAT \cite{DBLP:conf/iclr/VelickovicCCRLB18}.
However, most of the existing GNNs can only access the label information at the output when learning the mapping function between node representations and labels via \equref{equ:cross_entropy_loss}, corresponding to \ding{173} in \figref{fig:label_usage}.

\subsection{Combining Label Information with GNNs}
For better usage of labels, some recent methods additionally leverage the labels to augment node features \cite{DBLP:journals/corr/abs-2103-13355,DBLP:conf/ijcai/ShiHFZWS21} or optimize graph structure \cite{DBLP:conf/ijcai/0002KCJ0G19,wang2021combining,chen2019highwaygraph,DBLP:conf/ijcnn/YangYDCC21} at the input via \ding{174} in \figref{fig:label_usage}. 
On the one hand, the augmentation of node features could be denoted by rewriting \equref{equ:input_gnn} as
\begin{equation}
\bm{H}^0 = Augment(\bm{X}, \bm{Y}),
\end{equation}
where $Augment(\bm{X}, \bm{Y})= \bm{X} \| \bm{Y}$ represents the concatenation in \citet{DBLP:journals/corr/abs-2103-13355}. $Augment(\bm{X}, \bm{Y})= \bm{X} + \bm{Y} \bm{W}_{L}$ denotes the addition with transformed matrix $\bm{W}_{L} \in \mathbb{R}^{C \times F}$ for node label matrix $\bm{Y}$ in \citet{DBLP:conf/ijcai/ShiHFZWS21}. On the other hand, the optimization of graph structure is formulated as
\begin{equation}\label{equ:structure_optimization}
\bm{A}^\prime = Optimize(\bm{A}, \bm{Y}).
\end{equation}
In \citet{chen2019highwaygraph,DBLP:conf/ijcnn/YangYDCC21}, $Optimize(\bm{A}, \bm{Y}) = \bm{A} \cup \bm{S}$, where $\cup$ denotes the element-wise OR logical operation. $\bm{S} \in \mathbb{R}^{M \times M}$ establishes the connection between two nodes if they share the same label. In particular, $S_{ij}$ is set to 1 if $\bm{Y}_i$ is identical to $\bm{Y}_j$, and 0 otherwise.
$Optimize(\bm{A}, \bm{Y})$ is achieved by using LPA to reweight the edges based on the assumption that connected nodes should have the same label in \citet{DBLP:conf/ijcai/0002KCJ0G19,wang2021combining}. 
Then, \citet{chen2019highwaygraph,DBLP:conf/ijcnn/YangYDCC21,DBLP:conf/ijcai/0002KCJ0G19,wang2021combining} replace the $\overline{\bm{A}}$ in \equref{equ:convolution_gnn} with $\bm{A}^\prime$ and perform graph convolutions by
\begin{equation}
\bm{H}^{k+1}  = \sigma\left(\bm{A}^\prime \bm{H}^k\bm{W}^k\right).
\end{equation}
We conclude that \citet{chen2019highwaygraph,DBLP:conf/ijcnn/YangYDCC21,DBLP:conf/ijcai/0002KCJ0G19,wang2021combining} can optimize the original graph structure $\bm{A}$ to be more suitable for node classification and facilitate the task. However, although the above approaches provide insightful solutions for using labels, they still fail to capture the abundant information of labels (discussed in \secref{section-1}).

\begin{figure}[!htbp]
    \centering
    \subfigure[Traditional GNNs.]{
        \includegraphics[width=.42\columnwidth]{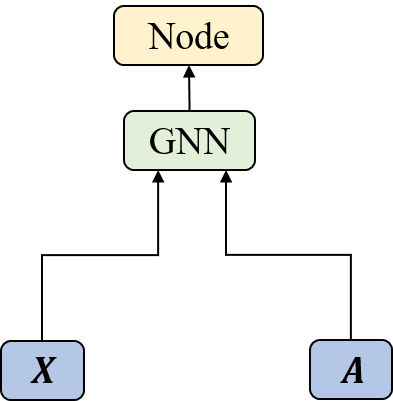}
        \label{fig:traditional_GNN}
    }
    \subfigure[Feature-augmented GNNs.]{
        \includegraphics[width=.42\columnwidth]{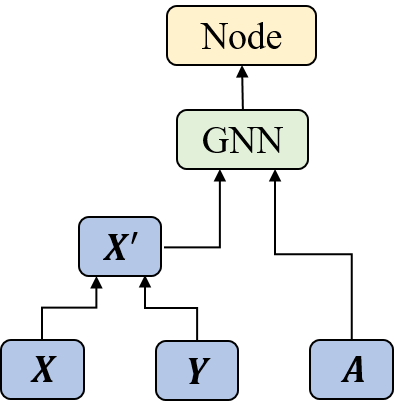}
        \label{fig:feature_augment_GNN}
    }
    \subfigure[Structure-optimized GNNs.]{
        \includegraphics[width=.42\columnwidth]{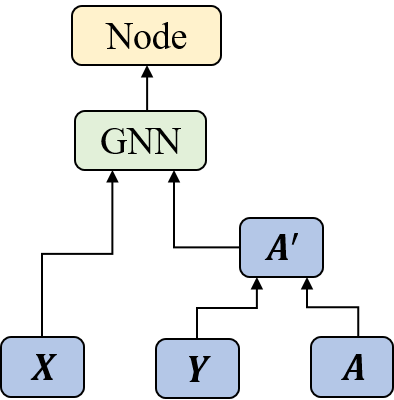}
        \label{fig:structure_optimize_GNN}
    }
    \subfigure[Label-enhanced GNNs.]{
        \includegraphics[width=.42\columnwidth]{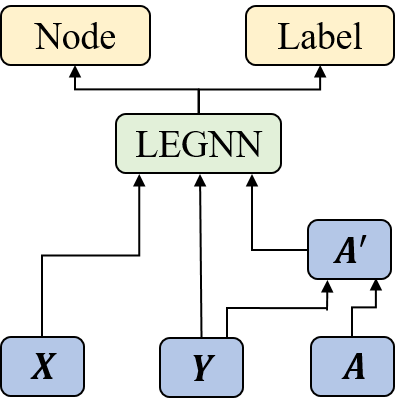}
        \label{fig:label_enhanced_GNN}
    }
    \caption{Comparisons of the existing methods with our approach.}
    \label{fig:different_GNNs}
\end{figure}

The learning paradigms of traditional GNNs and GNNs combined with label information are shown in \figref{fig:different_GNNs} (a), (b) and (c). Different from these methods, in this paper, we propose a label-enhanced semi-supervised learning framework to integrate rich label information into GNNs, which could jointly learn the representations of both nodes and labels, see \figref{fig:different_GNNs} (d).

\subsection{Self-Training on GNNs}
Recent studies have attempted to apply the self-training strategy on graphs, which add unlabeled nodes to the training set with pseudo labels \cite{DBLP:conf/aaai/LiHW18,zhou2019effective,DBLP:conf/ijcnn/YangYDCC21,DBLP:conf/aaai/SunLZ20,sun2021scalable}. As a self-training method on GNNs, \citet{DBLP:conf/aaai/LiHW18} first trained a GNN with the set of labeled nodes $\mathcal{L}$, and then added the most confident unlabeled nodes $\mathcal{U}^\prime \subseteq \mathcal{U}$ to $\mathcal{L}$ to obtain an enlarged training set $\mathcal{L} \ \cup \ \mathcal{U}^\prime$. \citet{zhou2019effective,sun2021scalable} selected pseudo-labeled nodes according to a pre-defined threshold, indicating that $\mathcal{U}^\prime$ only consists of unlabeled nodes whose predicted probabilities are higher than the threshold. 
\citet{DBLP:conf/ijcnn/YangYDCC21} utilized multiple GNNs to make predictions and assigned pseudo labels to the nodes whose predictions are identical among all the GNNs. \citet{DBLP:conf/aaai/SunLZ20} designed a multi-stage self-supervised training algorithm with a DeepCluster \cite{DBLP:conf/eccv/CaronBJD18} self-checking mechanism to select more precise pseudo labels. The objective function of self-training methods could be formulated as follows,
\begin{equation}
    \label{equ:joint_ce_pseudo_loss}
    L = \frac{1}{|\mathcal{L}|} \sum_{v_i \in \mathcal{L}} L(\bm{Y}_i, \hat{\bm{Y}}_i) + \frac{\lambda}{|\mathcal{U}^\prime|} \sum_{v_i \in \mathcal{U}^\prime} L(\widetilde{\bm{Y}}_i, \hat{\bm{Y}}_i)
\end{equation}
where $\widetilde{\bm{Y}}_{i}$ denotes the pseudo label for unlabeled node $v_i$. $\lambda$ is a hyperparameter to control the weight of pseudo labels.

However, these designs may make models sensitive to the initialization of parameters and affect the reliability of pseudo labels. In this paper, we design an adaptive self-training strategy to provide more reliable pseudo labels, and distinguish the importance of each pseudo-labeled node.

\section{Methodology}
\label{section-4}

This section first introduces the framework of the proposed model and then presents each component.

\begin{figure*}[!ht]
    \centering
    \includegraphics[width=1.75\columnwidth]{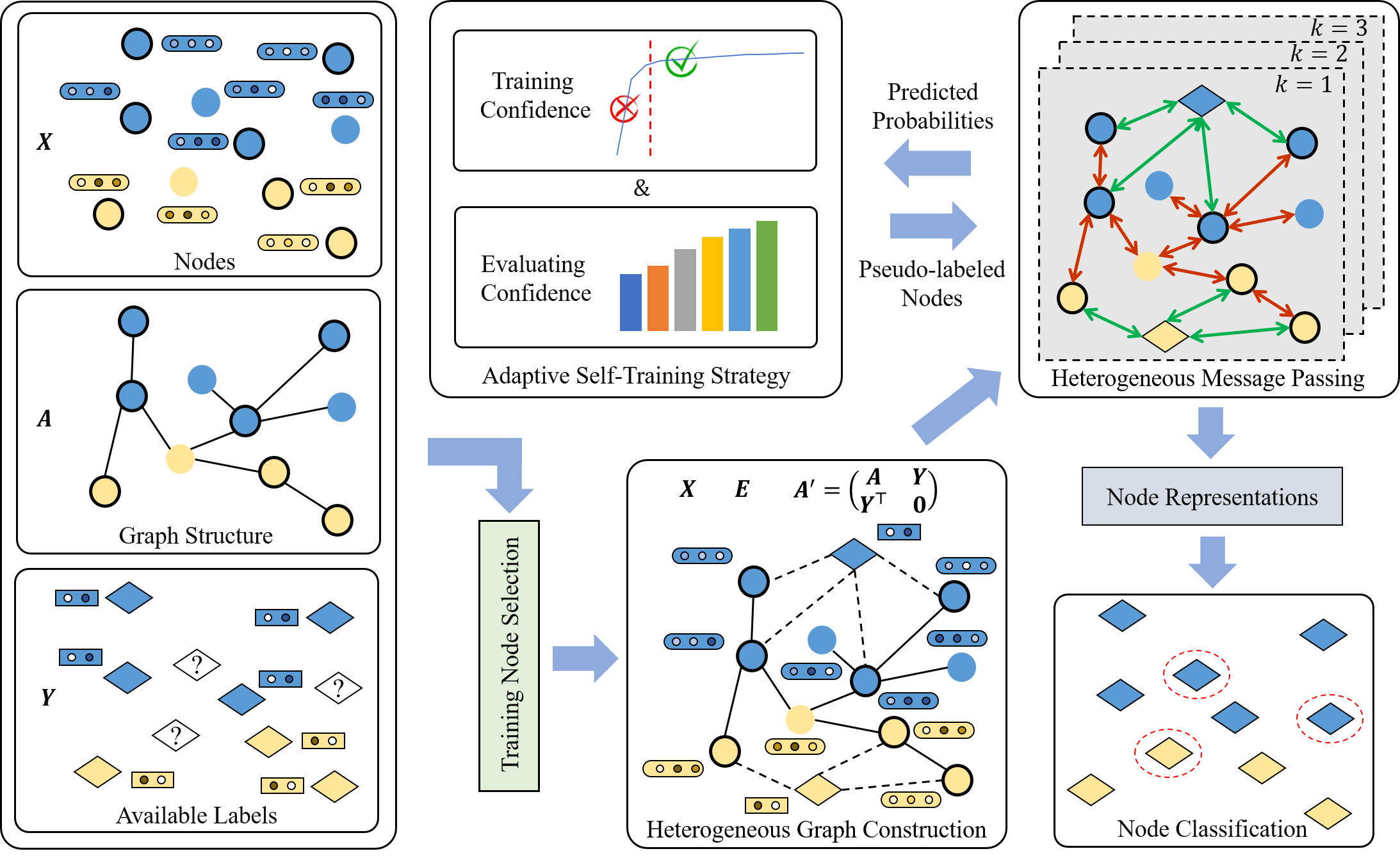}
\caption{Framework of the proposed approach.}
\label{fig:framework}
\end{figure*}

\subsection{Overview of the Proposed Framework}
\figref{fig:framework} illustrates our framework. We propose a Label-Enhanced Graph Neural Network (LEGNN) to comprehensively integrate rich label information into GNNs. In particular, LEGNN first constructs a heterogeneous graph based on the nodes, graph structure and available labels.
Then, it performs heterogeneous message passing on the constructed graph to jointly learn representations of both nodes and labels. A training node selection technique is presented to tackle the potential label leakage issue and guarantee the model generalization ability. We also design an Adaptive Self-Training strategy (AS-Train) to iteratively enlarge the training set with more reliable pseudo labels and distinguish the importance of each pseudo-labeled node according to the training and evaluating confidence.
During the training procedure, LEGNN provides the predicted probabilities of unlabeled nodes to AS-Train and AS-Train supplies reliable pseudo-labeled nodes for LEGNN.
Finally, the learned node representations are used to classify unlabeled nodes.

\subsection{Label-Enhanced Graph Neural Network}
LEGNN integrates the rich label information into GNNs via: 1) Heterogeneous Graph Construction; 2) Heterogeneous Message Passing; and 3) Training Node Selection.

\textbf{Heterogeneous Graph Construction}. 
We construct a heterogeneous graph to involve the information of both nodes and labels, due to its power in representing diverse properties \cite{DBLP:journals/sigkdd/SunH12,DBLP:journals/tkde/ShiLZSY17}. Formally, given a graph $\mathcal{G}$ that contains $M$ nodes with node feature matrix $\bm{X} \in \mathbb{R}^{M \times F}$ and adjacency matrix $\bm{A} \in \mathbb{R}^{M \times M}$, we first inject $C$ labels as a new type of nodes into the graph with label feature matrix $\bm{E} \in \mathbb{R}^{C \times F^\prime}$, where $F^\prime$ is the number of label features. Then, a node is connected with its corresponding label if it belongs to the labeled set $\mathcal{L}$. We can find that the adjacency matrix between nodes and labels is exactly identical to the node label matrix $\bm{Y} \in \mathbb{R}^{M \times C}$. Finally, the heterogeneous graph $\mathcal{G}^\prime$ is obtained, including the node feature matrix $\bm{X}$, label feature matrix $\bm{E}$, and adjacency matrix $\bm{A}^\prime \in \mathbb{R}^{(M+C) \times (M+C)}$, which is denoted by
\begin{equation}\label{equ:construct_graph}
\bm{A}^\prime = \begin{pmatrix} \bm{A} & \bm{Y} \\ \bm{Y}^\top & \bm{0} \end{pmatrix}.
\end{equation}
In this work, we use the one-hot encodings of labels to represent $\bm{E}$ and leverage $\bm{0} \in \mathbb{R}^{C \times C}$ to denote that labels are not directly connected with each other. But both of them can be enriched when the relevant prior knowledge of label features or label correlations is explicitly given. If we further perform message passing on $\mathcal{G}^\prime$, labels can contribute in two aspects. Firstly, each label serves as a virtual center for intra-class nodes and makes them become 2-hop neighbors even if they are distant from each other in the original $\mathcal{G}$. This provides the possibility to enhance the smoothness of intra-class node representations, which is highly desired for classifying nodes. Secondly, label semantics are modelled via $\bm{E}$, which is helpful to discover the semantic correlations of labels. Although there are no direct connections between labels, they could still receive messages from each other via high-order interactions, which would help mine their implicit relationships.

\textbf{Heterogeneous Message Passing}. To learn on the heterogeneous graph $\mathcal{G}^\prime$, we perform heterogeneous message passing by designing different parameters for nodes and labels, respectively. We first align the feature dimension of node feature matrix $\bm{X} \in \mathbb{R}^{M \times F}$ and label feature matrix $\bm{E} \in \mathbb{R}^{C \times F^\prime}$ by
\begin{equation}\label{equ:align_dimension}
\begin{aligned}
\bm{H}_N^0 &= \bm{X}\bm{P}_N, \\
\bm{H}_L^0 &= \bm{E} \bm{P}_L,
\end{aligned}
\end{equation}
where $\bm{P}_N \in \mathbb{R}^{F \times D}$ and $\bm{P}_L \in \mathbb{R}^{F^\prime \times D}$ are the projection matrices specific to nodes and labels. $D$ denotes the hidden dimension. The projected inputs can be denoted by rewriting \equref{equ:input_gnn} as
\begin{equation}\label{equ:input_feature}
\bm{H}^0 = \begin{pmatrix} \bm{H}_N^0 \\ \bm{H}_L^0 \end{pmatrix} \in \mathbb{R}^{(M+C) \times D}.
\end{equation}
Then, we extend \equref{equ:convolution_gnn} to support heterogeneous message passing as follows,
\begin{equation}\label{equ:heterogeneous_message_passing}
\begin{aligned}
\bm{H}^{k+1} &= \begin{pmatrix} \bm{H}_N^{k+1} \\ \bm{H}_L^{k+1} \end{pmatrix} = \sigma\left(\bm{A}^\prime \bm{H}^{k} \bm{W}^{k}\right) \\
&=\sigma\left(\begin{pmatrix} \bm{A} & \bm{Y} \\ \bm{Y}^\top & \bm{0} \end{pmatrix} \begin{pmatrix} \bm{H}_N^{k} \bm{W}_N^{k}\\ \bm{H}_L^{k} \bm{W}_L^{k} \end{pmatrix} \right),
\end{aligned}
\end{equation}
where $\bm{W}_N^{k} \in \mathbb{R}^{D \times D}$ and $\bm{W}_L^{k} \in \mathbb{R}^{D \times D}$ are the transformation matrices for nodes and labels at the $k$-th layer. Specifically, representations of nodes and labels are computed by
\begin{equation}\label{equ:separate_heterogeneous_message_passing}
\begin{aligned}
\bm{H}_N^{k+1} = \sigma &\left(\bm{A} \bm{H}_N^k \bm{W}_N^k + \bm{Y} \bm{H}_L^k \bm{W}_L^k\right), \\
\bm{H}_L^{k+1} &= \sigma\left(\bm{Y}^\top \bm{H}_N^k \bm{W}_N^k\right).
\end{aligned}
\end{equation}
Finally, we could obtain both node representations $\bm{Z}_N$ and label representations $\bm{Z}_L$ by rewriting \equref{equ:output_gnn} as
\begin{equation}\label{equ:output_feature}
\bm{Z} = \begin{pmatrix} \bm{Z}_N \\ \bm{Z}_L \end{pmatrix} = \begin{pmatrix} \bm{H}_N^{K} \\ \bm{H}_L^{K} \end{pmatrix}.
\end{equation}

A benefit of the heterogeneous message passing is that it is \textit{applicable to any message passing GNN}. When applying to various GNNs, we just need to additionally employ specialized trainable parameters for labels. The differences of various GNNs mainly lie in the calculation of adjacency matrix $\bm{A}^\prime$ in \equref{equ:heterogeneous_message_passing}, such as the normalized, sampling-based and attention-based adjacency matrices in GCN, GraphSAGE and GAT, respectively. Take the GAT as an example, the weight of edge $e=(u,v)$ at layer $k$ is calculated by,
\begin{equation}
    \beta_{u,v}^k = \xi\left(\left[{\bm{a}_{\phi(u)}^k \| \bm{a}_{\phi(v)}^k}\right]^\top \left[\bm{W}_{\phi(u)}^k \bm{h}_{u}^{k-1} \| \bm{W}_{\phi(v)}^k \bm{h}_{v}^{k-1} \right]\right), \notag
\end{equation}
\begin{equation}
    A_{u,v}^{\prime k} = \frac{\exp{(\beta_{u,v}^k)}}{\sum_{v\prime \in \mathcal{N}_u} \exp{(\beta_{u,v\prime}^k)}},
\end{equation}
where $\xi(\cdot)$ is the LeakyReLU activation function. $\mathcal{N}_u$ is the set of node $u$'s neighbors. $\phi(u)$ is the type mapping function that maps nodes to type $N$ or type $L$. $\bm{a}_{\phi(u)}^k$ denotes the attention vector for nodes with type $\phi(u)$ at layer $k$.

Our approach is different from the existing heterogeneous graph learning models (e.g., \cite{DBLP:conf/www/WangJSWYCY19,DBLP:conf/kdd/ZhangSHSC19,DBLP:conf/aaai/ZhaoWSHSY21,ji2021heterogeneous}) because LEGNN performs message passing on the constructed heterogeneous graph with nodes and labels, and aims to incorporate the label information into the learning process of GNNs, while existing methods ignore the role of labels and mainly focus on handling the types of different nodes.

\textbf{Training Node Selection}.
When constructing the heterogeneous graph $\mathcal{G}^\prime$, if we connect all the training nodes with their labels, the model will trivially predict their labels by solely using the connecting information and lead to poor generalization ability as testing nodes are not connected with labels (which is validated by the experiments in \secref{section-5-training-node-selection}). Therefore, we present a training node selection technique to cope with the potential label leakage issue and guarantee model generalization ability. 

Specifically, at each training epoch, we first randomly select a subset of labeled nodes $\widetilde{\mathcal{L}}$ according to a pre-defined training node selection rate $\alpha$, s.t. $|\widetilde{\mathcal{L}}| = \lfloor \alpha * |\mathcal{L}| \rfloor $, whose labels are used as the ground truth to train model. Then, we establish the connections between the remaining nodes in $\mathcal{L} \backslash \widetilde{\mathcal{L}}$ and their labels to construct the heterogeneous graph $\mathcal{G}^\prime$. As a result, each labeled node is either used as the ground truth or used to construct the heterogeneous graph, but \textit{none of them will do both}. Therefore, the prediction task is nontrivial and could guide our model to learn how to predict the label of a node from its own features and the neighboring nodes’ features and labels. Moreover, predicting the labels of nodes in $\widetilde{\mathcal{L}}$ can simulate the predictions of unlabeled nodes in $\mathcal{U}$, and the generalization ability of our model is guaranteed. Hence, the label leakage issue is solved by our training node selection mechanism. \textit{During the inference process, we will connect all the training nodes with their labels and predict the unlabeled nodes.}

\subsection{Adaptive Self-Training Strategy}
Existing self-training methods usually add unlabeled nodes to the training set with pseudo labels if they are the most confident \cite{DBLP:conf/aaai/LiHW18,DBLP:conf/aaai/SunLZ20} or their predicted probabilities are higher than a pre-defined threshold \cite{zhou2019effective,DBLP:conf/ijcnn/YangYDCC21,sun2021scalable}. However, these designs ignore the reliability of models during the training process and further affect the quality of pseudo labels. 
Moreover, though these methods control the weight of pseudo labels via $\lambda$ in \equref{equ:joint_ce_pseudo_loss}, they fail to differentiate the importance inside the pseudo-labeled nodes. In this paper, we propose an adaptive self-training strategy to provide more reliable pseudo labels using the training confidence, and distinguish the importance of each pseudo-labeled node by the evaluating confidence. We illustrate our motivation by the empirical analysis on the ogbn-arxiv dataset (which will be introduced in \secref{section-5}).

\begin{figure}[!ht]
    \centering
    \subfigure[Classification accuracy during the training process.]{
        \includegraphics[width=.47\columnwidth]{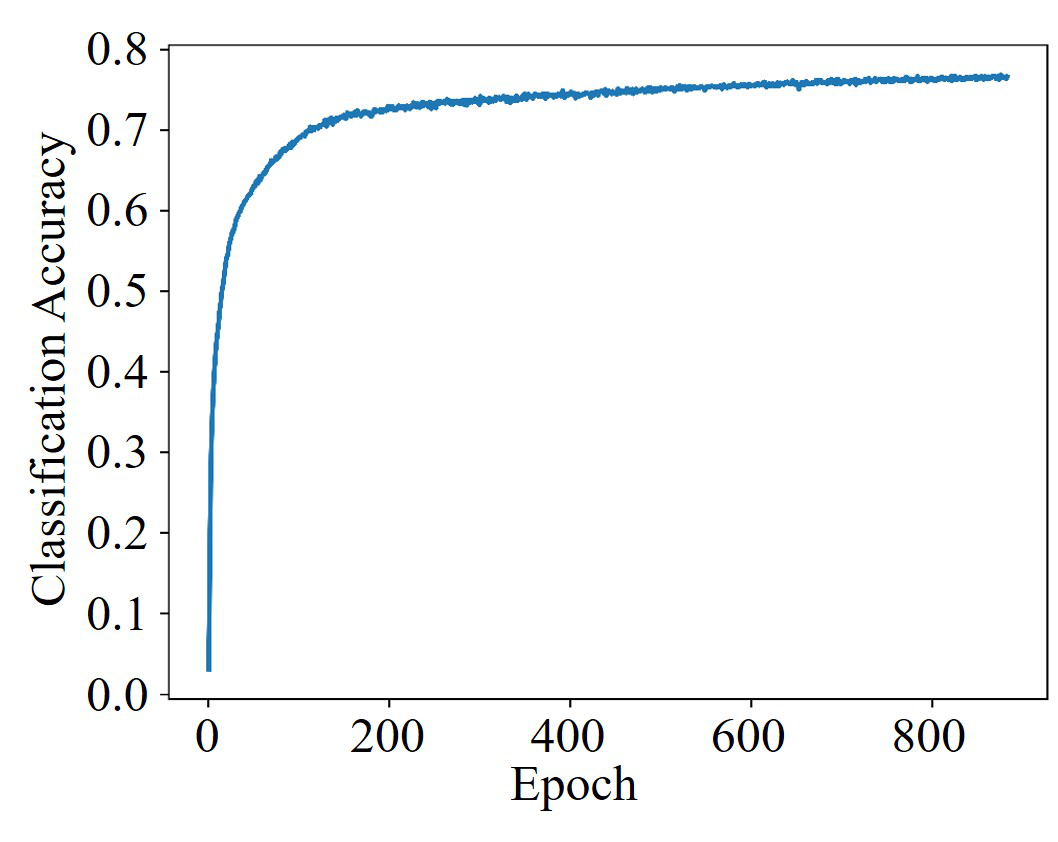}
        \label{fig:train_accuracy_arxiv}
    }
    \subfigure[Approximation of the accuracy convergence curve.]{
        \includegraphics[width=.47\columnwidth]{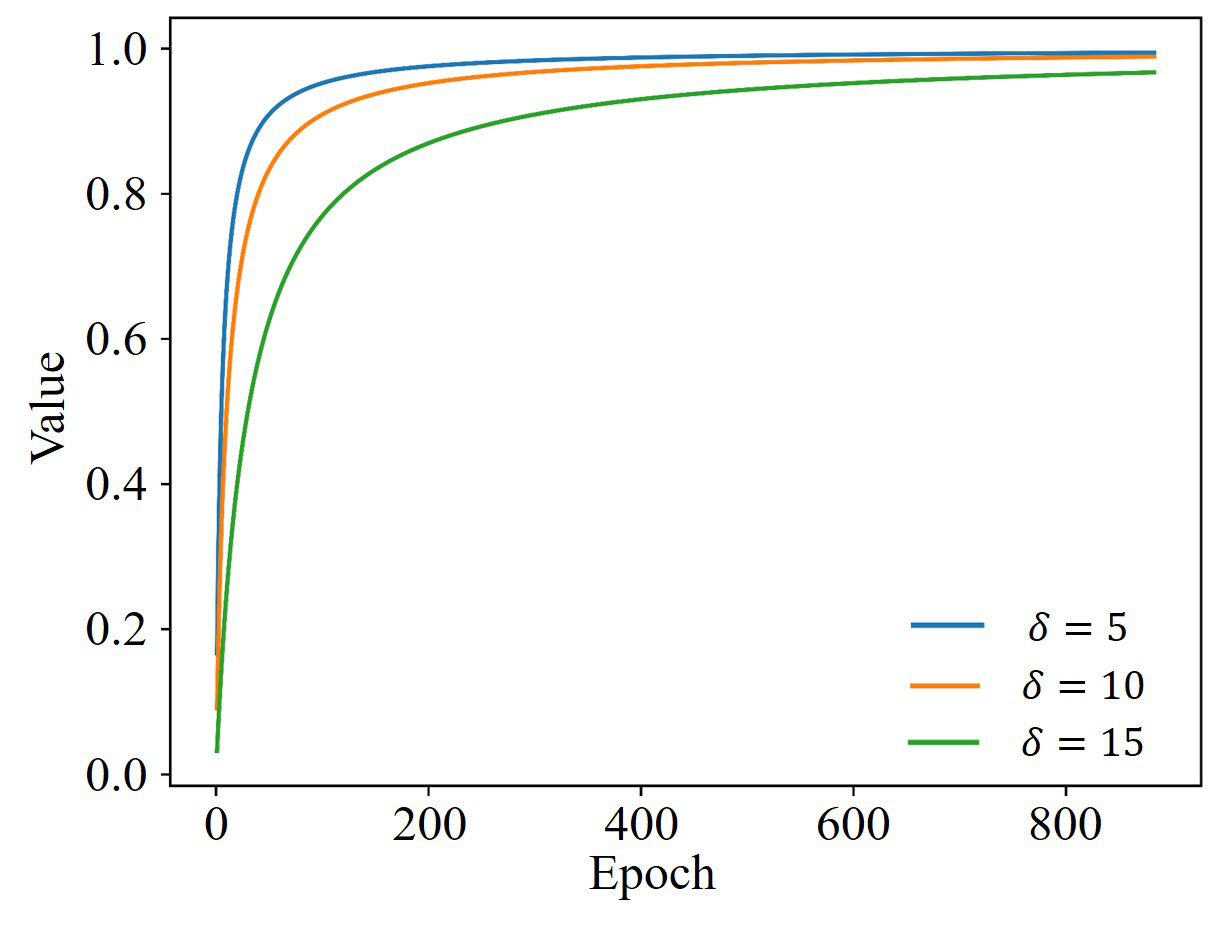}
        \label{fig:different_temperature_log}
    }
    \subfigure[Accuracy of pseudo labels w/ and w/o the training confidence.]{
        \includegraphics[width=.47\columnwidth]{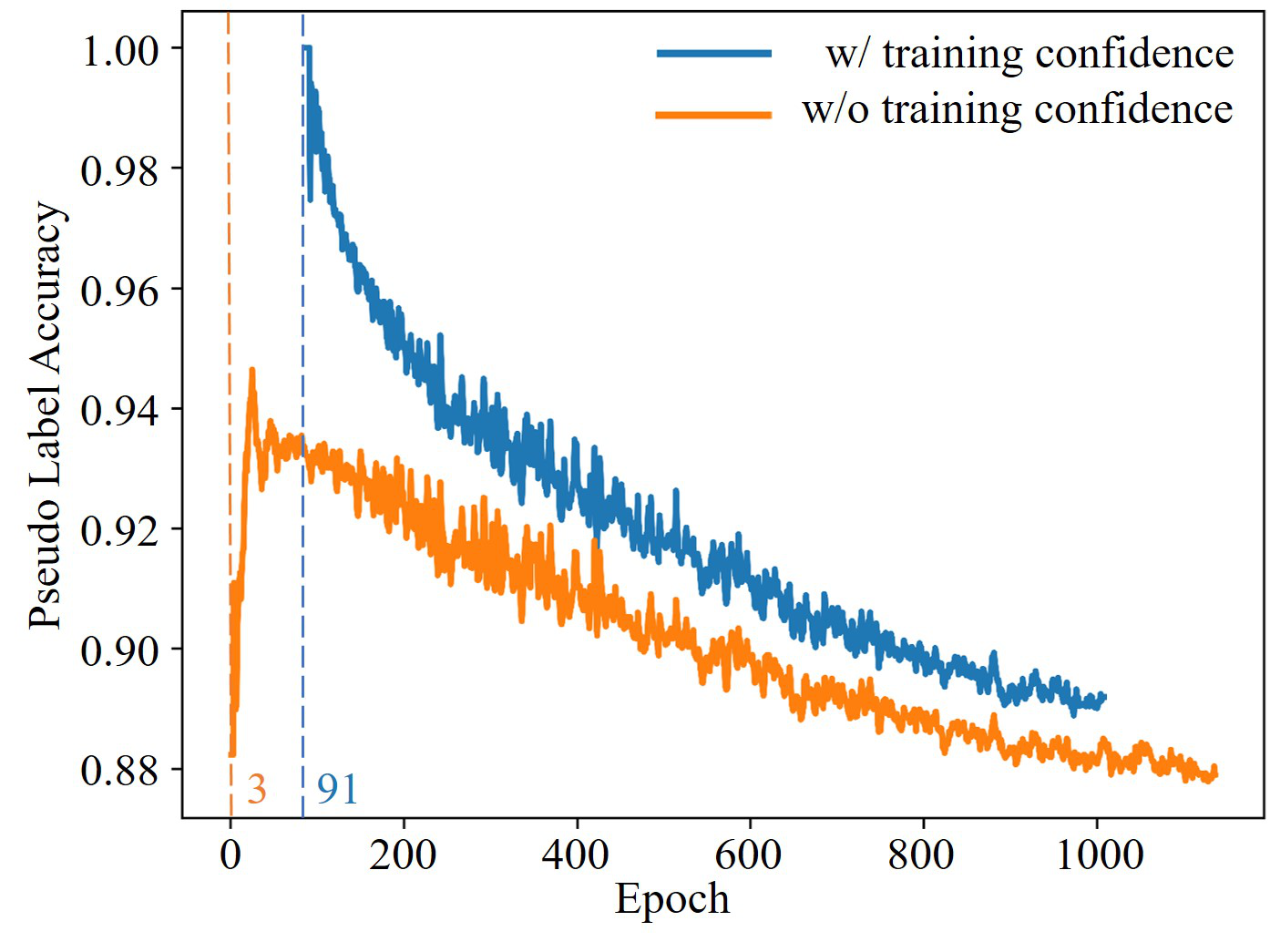}
        \label{fig:train_w_o_train_confidence}
    }
    \subfigure[Accuracy of pseudo labels with different probabilities.]{
        \includegraphics[width=.47\columnwidth]{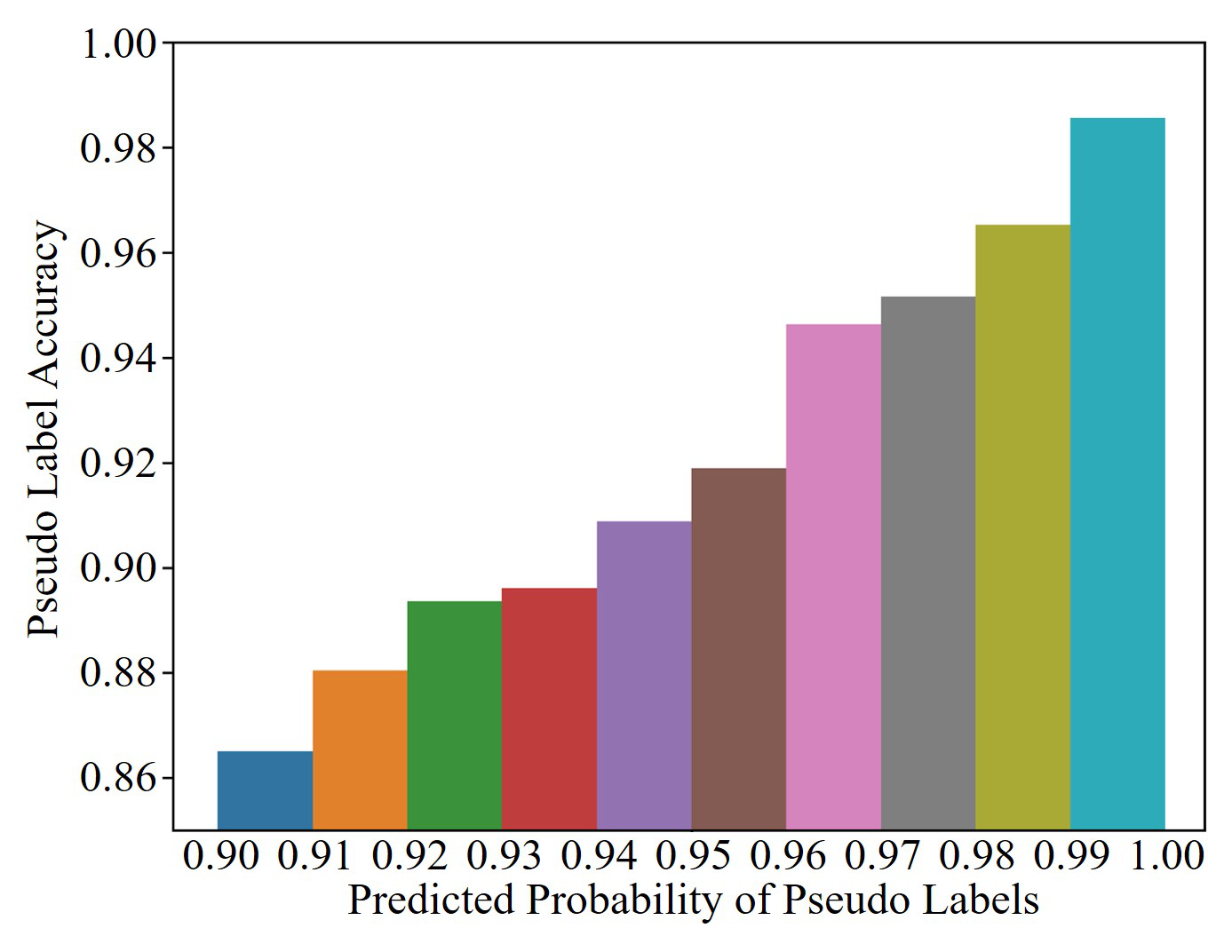}
        \label{fig:pseudo_label_accuracy}
    }
    \caption{Illustration of the adaptive self-training strategy on the ogbn-arxiv dataset.}
    \label{fig:self_training}
\end{figure}

\textbf{Training Confidence}.
We assume that pseudo labels given by a model are reliable only when the model can well fit the training set at least. As shown in \figref{fig:train_accuracy_arxiv}, we estimate the model ability in fitting the training set by the classification accuracy. 
To overcome the varied ranges of accuracy on different datasets, we present the training confidence ($TC$) to approximate the accuracy convergence curve as follows,
\begin{equation}\label{equ:training_confidence}
    TC = sigmoid\left(\log \left( e / \delta \right) \right),
\end{equation}
where $sigmoid(\cdot)$ activation function is used to constrain $TC$ between 0 and 1. $e$ denotes the current training epoch. $\delta$ is a scale factor to control the convergence speed. Curves with different $\delta$ are shown in \figref{fig:different_temperature_log}, indicating that we should set appropriate $\delta$ for various datasets to approximate the accuracy convergence curve.
When judging whether an unlabeled node $v_i$ should be assigned with a pseudo label, we first calculate $v_i$'s predicted probability $p_i$, and then multiply it with the training confidence $TC$. If the product is higher than the pre-defined threshold $t$, it indicates that $v_i$ is more likely to be classified correctly, and then $v_i$ would be added to the training set with pseudo label $\widetilde{y}_i$ as follows,
\begin{equation}\label{equ:pseudo_node_predict_probability_and_label}
\begin{aligned}
& \quad \quad p_i = \mathop{\max}_{1 \leq c \leq C} \hat{Y}_{ic}, \\
\widetilde{y}_i = & \mathop{\arg\max}_{1 \leq c \leq C}\hat{Y}_{ic} \ \text{ if } \ p_i * TC > t.
\end{aligned}
\end{equation}

\figref{fig:train_w_o_train_confidence} shows the pseudo label accuracy of using and without using the training confidence. We observe that the training confidence can: 1) consistently and significantly improve the accuracy of pseudo labels; 2) postpone the generation of pseudo nodes until the model can well fit the training set (starts to generate pseudo labels at the 91-th instead of the 3-rd epoch). 
Note that with the increase of training epochs, the number of pseudo-labeled nodes becomes larger. Classifying more and more nodes is rather tough so the accuracy of pseudo labels decreases gradually, but the number of correct pseudo labels still keeps rising.

\textbf{Evaluating Confidence}. 
With the assumption that pseudo labels with higher predicted probabilities are more reliable than those with lower predicted probabilities, we introduce the Evaluating Confidence ($EC$) to distinguish the importance of each pseudo-labeled node.

\figref{fig:pseudo_label_accuracy} shows the correlations between predicted probability and accuracy of pseudo labels. We observe that though all the predicted probabilities are more than the pre-defined threshold $t$ (i.e., 0.9 in this analysis), higher predicted probabilities correspond to more accurate pseudo labels, e.g., 0.9857 with probability in (0.99, 1.00] vs. 0.9089 with probability in (0.94, 0.95]. This indicates that pseudo labels with higher predicted probabilities are more reliable and should contribute more in the model optimization. Hence, we rewrite \equref{equ:joint_ce_pseudo_loss} as the objective function to consider each pseudo-labeled node's importance by
\begin{equation}\label{equ:joint_ce_pseudo_importance_loss}
    L = \frac{1}{|\mathcal{L}|} \sum_{v_i \in \mathcal{L}} L(\bm{Y}_i, \hat{\bm{Y}}_i) + \frac{\lambda \cdot TC}{|\mathcal{U}^\prime|} \sum_{v_i \in \mathcal{U}^\prime} {EC}_i \cdot L(\widetilde{\bm{Y}}_i, \hat{\bm{Y}}_i)
\end{equation}
where ${EC}_i$ is the evaluating confidence of unlabeled node $v_i$, and we use $p_i$ in \equref{equ:pseudo_node_predict_probability_and_label} to represent it. If ${EC}_i$ is higher, it means the pseudo label of node $v_i$ is more likely to be correct, and our model tends to focus more on node $v_i$ by minimizing $L(\widetilde{\bm{Y}}_i, \hat{\bm{Y}}_i)$, and vice versa. It is worth noticing that the pseudo-labeled nodes are relatively stable during the training process because once the model can confidently assign nodes with pseudo labels, these nodes are also likely to be included in subsequent epochs.

\subsection{Model Learning Process}
We obtain the predicted probability $\hat{\bm{Y}} \in \mathbb{R}^{M \times C}$ over all the $C$ label classes by
\begin{equation}\label{equ:classification}
    \hat{\bm{Y}} = softmax\left(\bm{Z}_N \bm{W}_{pred} + \bm{b}_{pred}\right),
\end{equation}
where $\bm{W}_{pred} \in \mathbb{R}^{D \times C}$ and $\bm{b}_{pred} \in \mathbb{R}^C$ are learnable parameters. Our model is trained with \equref{equ:joint_ce_pseudo_importance_loss} as the objective function and is optimized by the back propagation algorithm. The training process of the proposed LEGNN with AS-Train is shown in \algoref{alg:training_process}. 

\begin{algorithm}
\SetKwComment{Comment}{/* }{ */}
\SetKwInOut{Input}{Input}
\SetKwInOut{Output}{Output}
\caption{Training process of LEGNN with the AS-Train strategy}
\label{alg:training_process}
\Input{Graph $\mathcal{G}=\left(\mathcal{V},\mathcal{E}\right)$ with adjacency matrix $\bm{A}$, node feature matrix $\bm{X}$, node label matrix $\bm{Y}$, label feature matrix $\bm{E}$, set of labeled nodes $\mathcal{L} \subseteq \mathcal{V}$, hyperparameters $\alpha$, $\delta$, $t$ and $\lambda$, maximal training epochs $MaxEpoch$\;}
\Output{The model parameters $\Theta$ after training\;}
Initialize the parameters in LEGNN with random weights $\Theta$\;
Initialize $\mathcal{U}^\prime$ to be an empty set\;
$Epochs \gets 1$\;
\While{not converge and $Epochs \leq MaxEpoch$}{
    Get $\widetilde{\mathcal{L}}$ via the training node selection with $\mathcal{L}$ and $\alpha$ as inputs\;
    Construct the heterogeneous graph $\mathcal{G}^\prime$ with adjacency matrix $\bm{A}^\prime$ by connecting nodes in $\mathcal{L} \backslash \widetilde{\mathcal{L}}$ with their labels via \equref{equ:construct_graph} with $\bm{A}$ and $\bm{Y}$ as inputs\;
    $\hat{\bm{Y}} \gets$ Compute the predicted probabilities via \equref{equ:align_dimension}, \equref{equ:heterogeneous_message_passing}, \equref{equ:output_feature} and \equref{equ:classification} with $\bm{A}^\prime$, $\bm{X}$, $\bm{E}$ and $\Theta$ as inputs\;
    $\mathcal{U}^\prime$, $\widetilde{\bm{Y}}$, $TC$, $\{{EC}_i\} \gets$ Get the set of pseudo-labeled nodes, the pseudo labels, the training confidence and the evaluating confidence via \equref{equ:training_confidence} and \equref{equ:pseudo_node_predict_probability_and_label} with $\hat{\bm{Y}}$, $Epochs$, $\delta$ and $t$ as inputs\;
    Optimize the model parameters $\Theta$ by back propagation via \equref{equ:joint_ce_pseudo_importance_loss} with $\bm{Y}$, $\hat{\bm{Y}}$, $\widetilde{\bm{Y}}$, $\mathcal{L}$, $\mathcal{U}^\prime$, $TC$, $\{{EC}_i\}$ and $\lambda$ as inputs\;
    $Epochs \gets Epochs + 1$\;
}
\end{algorithm}

\section{Experiments}
\label{section-5}

In this section, we conduct extensive experiments on both the real-world and synthetic datasets to show the effectiveness of our approach and present detailed analysis. 

\subsection{Descriptions of Real-world Datasets}
We conduct experiments on three real-world datasets.

\begin{table*}[!htbp]
\centering
\caption{Statistics of the datasets.}
\label{tab:dataset}
\begin{tabular}{c|cccccc}
\hline
Datasets   & \#Nodes   & \#Edges   & \#Classes  & \multicolumn{1}{c}{Split Ratio (\%)} & Split Strategy & Homophily \\ \hline
ogbn-arxiv & 169,343   & 1,166,243  & 40 & 54 / 18 / 28       & Time-based Split \cite{DBLP:conf/nips/HuFZDRLCL20}           & 0.6551    \\
ogbn-mag   & 1,939,743 & 21,111,007 & 349 & 85 / 9 / 6       & Time-based Split  \cite{DBLP:conf/nips/HuFZDRLCL20}          & 0.3040    \\
oag-venue  & 731,050   & 3,642,689  & 241 & 64 / 15 / 21     & Time-based Split \cite{yu2022heterogeneous}             & 0.2324    \\ \hline
\end{tabular}
\end{table*}

\begin{itemize}
    \item \textbf{ogbn-arxiv} is a directed graph that represents the citation network between all Computer Science arXiv papers \cite{DBLP:conf/nips/HuFZDRLCL20}. Each node denotes an arXiv paper and each directed edge indicates that one paper cites another one. Each paper is associated with a 128-dimensional feature vector by averaging the embeddings of words in its title and abstract, which are computed by Word2Vec \cite{DBLP:conf/nips/MikolovSCCD13}. The task is to predict the subject areas of each paper. 
    
    \item \textbf{ogbn-mag} is a heterogeneous academic graph extracted from the Microsoft Academic Graph (MAG) \cite{DBLP:journals/qss/WangSHWDK20}, including papers, authors, fields and institutions, as well as the relations among nodes \cite{DBLP:conf/nips/HuFZDRLCL20}. Each paper is associated with a 128-dimensional Word2Vec feature. For nodes that do not have features, we generate their features by metapath2vec \cite{DBLP:conf/kdd/DongCS17}. The task is to predict the venue of each paper.
    
    \item \textbf{oag-venue} is a heterogeneous academic graph extracted from the Open Academic Graph (OAG) \cite{DBLP:conf/kdd/ZhangLTDYZGWSLW19}, consisting of papers, authors, fields, institutions and their connections \cite{yu2022heterogeneous}.
    The feature of each paper is a 768-dimensional vector, which is the weighted combination of each word's representation in the paper's title. The representation and attention score of each word are obtained from a pre-trained XLNet \cite{DBLP:conf/nips/YangDYCSL19}. The feature of each author is the average of his/her published paper features. Features of other types of nodes are generated by metapath2vec \cite{DBLP:conf/kdd/DongCS17}. The task is to predict the venue of each paper.
\end{itemize}
The task type on all the datasets is multi-class classification as each node belongs to exactly one label. All the datasets are split by the publication dates of papers and we follow the same data splits in \cite{DBLP:conf/nips/HuFZDRLCL20} and \cite{yu2022heterogeneous}. We define the homophily as the fraction of edges in a graph whose endpoints have the same label, that is, $\frac{|\left\{(u,v):(u,v) \in \mathcal{E} \wedge \bm{Y}_u=\bm{Y}_v\right\}|}{|\mathcal{E}|}$. For ogbn-mag and oag-venue, we calculate the graph homophily on the subgraph that only contains papers.
Statistics of the datasets are shown in \tabref{tab:dataset}. 

\subsection{Experimental Settings}
We train the models in a full-batch manner on ogbn-arxiv. 
Due to the memory limitation, we adopt a neighbor sampling strategy to train models in a mini-batch manner on ogbn-mag and oag-venue.
In particular, for each target node, we sample a fixed number of neighbors at each layer uniformly and then aggregate messages from the sampled neighbors layer by layer. 
Since LEGNN is stacked with $K=3$ graph convolutional layers in the experiments, we set the numbers of sampled neighbors to 15, 10, 5 in the first, second and third layer, respectively.
We transform ogbn-mag and oag-venue to homogeneous graphs by ignoring the types of nodes and relations to eliminate the effect of graph heterogeneity. Adam \cite{kingma2014adam} is used as the optimizer with cosine annealing learning rate scheduler \cite{DBLP:conf/iclr/LoshchilovH17}. We use dropout \cite{DBLP:journals/jmlr/SrivastavaHKSS14} to prevent over-fitting. Residual connections \cite{DBLP:conf/cvpr/HeZRS16} are employed for each graph convolutional layer. Batch normalization \cite{DBLP:conf/icml/IoffeS15} is applied on the ogbn-arxiv dataset. We train the models with 2000, 500, and 1000 epochs on ogbn-arxiv, ogbn-mag and oag-venue, respectively. An early stopping strategy is adopted with the patience of 200, 50 and 100.
The hidden dimensions of both nodes and labels are set to 540, 512 and 256 on ogbn-arxiv, ogbn-mag and oag-venue, respectively. 
We apply grid search to find the best settings.
The dropout and learning rate are searched in $\left[0.1, 0.2, 0.3, 0.4, 0.5, 0.6\right]$ and $\left[0.001, 0.002, 0.005\right]$ on ogbn-arxiv. On ogbn-mag and oag-venue, we search the dropout and learning rate in $\left[0.1, 0.2, 0.3, 0.4\right]$ and $\left[0.001, 0.01\right]$. The settings of dropout and learning rate of baselines and our method are shown in \tabref{tab:dropout_learning_rate} in \appendixref{section-appendix-hyper-parameters}. 
For hyperparameters, we set $\alpha$ to 0.5 on all the datasets. $\delta$ is chosen by approximating the accuracy convergence curve on each dataset. The searched ranges of $t$ and $\lambda$ are $\left[0.5, 0.6, 0.7, 0.8, 0.9\right]$ and $\left[0.05, 0.1, 0.3, 0.5, 1.0\right]$.
The hyperparameter settings of our approach are shown in \tabref{tab:hyperparameter_settings} in \appendixref{section-appendix-hyper-parameters}. The model with the highest accuracy on the validation set is used for testing.

We implement our approach based on PyTorch \cite{DBLP:conf/nips/PaszkeGMLBCKLGA19} and Deep Graph Library (DGL) \cite{DBLP:journals/corr/abs-1909-01315}. We run each model for multiple times with different seeds and report the average performance. The experiments are conducted on an Ubuntu machine equipped with one Intel(R) Xeon(R) Gold 6130 CPU @ 2.10GHz with 16 physical cores. 
The GPU device is NVIDIA Tesla T4 with 15 GB memory. The codes and datasets are available at https://github.com/yule-BUAA/LEGNN.

\subsection{Comparisons with SOTAs in Using Labels}
We validate the superiority of LEGNN in using labels by comparing it with state-of-the-arts that also leverage label information for GNNs. GCN \citet{DBLP:conf/iclr/KipfW17}, GraphSAGE \cite{DBLP:conf/nips/HamiltonYL17} and GAT \cite{DBLP:conf/iclr/VelickovicCCRLB18} are employed as the backbones. The compared methods include \textbf{Vanilla} GNNs, \textbf{Concat} \cite{DBLP:journals/corr/abs-2103-13355} and \textbf{Addition} \cite{DBLP:conf/ijcai/ShiHFZWS21}.
We report the results in \tabref{tab:label_usage}.
Although methods in \citet{chen2019highwaygraph,DBLP:conf/ijcai/0002KCJ0G19,wang2021combining,DBLP:conf/ijcnn/YangYDCC21} also use labels, we do not compare with them in the experiments because they are infeasible to be applied on large-scale datasets.
For methods in \citet{chen2019highwaygraph,DBLP:conf/ijcnn/YangYDCC21}, establishing the connections between each pair of nodes that belonging to the same label would make the graph too dense (introducing $\mathcal{O}(M^2)$ new edges) and the dense graph is infeasible to be loaded on the device. In \citet{DBLP:conf/ijcai/0002KCJ0G19,wang2021combining}, the objective functions are computed on the whole graph adjacency matrix, making them only feasible on small-scale datasets. We try to run the methods in \citet{chen2019highwaygraph,DBLP:conf/ijcai/0002KCJ0G19,wang2021combining,DBLP:conf/ijcnn/YangYDCC21} by using their official codes or our implementations, but all of them raise the out-of-memory (OOM) error even when running on the relatively small ogbn-arxiv dataset. 

\begin{table*}[!htbp]
\centering
\caption{Performance of different methods in using labels.}
\label{tab:label_usage}
\setlength{\tabcolsep}{1.2mm}{
\begin{tabular}{c|c|cccccc}
\hline
\multirow{3}{*}{Backbones} & \multirow{3}{*}{Methods} & \multicolumn{6}{c}{Datasets}                                                                                                                                                                                         \\ \cline{3-8} 
                           &                          & \multicolumn{2}{c|}{ogbn-arxiv}                                          & \multicolumn{2}{c|}{ogbn-mag}                                            & \multicolumn{2}{c}{oag-venue}                                  \\ \cline{3-8} 
                           &                          & Accuracy $\uparrow$                & \multicolumn{1}{c|}{Macro-F1 $\uparrow$}                 & Accuracy $\uparrow$                 & \multicolumn{1}{c|}{Macro-F1 $\uparrow$}                 & Accuracy $\uparrow$                           & Macro-F1 $\uparrow$                \\ \hline
\multirow{4}{*}{GCN}       & Vanilla                  & 0.7267 $\pm$ 0.0024          & \multicolumn{1}{c|}{0.5327 $\pm$ 0.0040}          & 0.4189 $\pm$ 0.0037          & \multicolumn{1}{c|}{0.2284 $\pm$ 0.0019}          & 0.2110 $\pm$ 0.0013                     & 0.1636 $\pm$ 0.0034          \\
                           & Concat                   & 0.7305 $\pm$ 0.0009          & \multicolumn{1}{c|}{0.5334 $\pm$ 0.0009}          & 0.4309 $\pm$ 0.0043           & \multicolumn{1}{c|}{0.2627 $\pm$ 0.0015}          & 0.2290 $\pm$ 0.0012                     & 0.1764 $\pm$ 0.0013          \\
                           & Addition                 & 0.7294 $\pm$ 0.0033          & \multicolumn{1}{c|}{0.5354 $\pm$ 0.0027}          & 0.4342 $\pm$ 0.0023           & \multicolumn{1}{c|}{0.2631 $\pm$ 0.0054}          & 0.2289 $\pm$ 0.0013                     & 0.1817 $\pm$ 0.0033          \\
                           & LEGNN                    & \textbf{0.7329 $\pm$ 0.0026} & \multicolumn{1}{c|}{\textbf{0.5361 $\pm$ 0.0029}} & \textbf{0.4710 $\pm$ 0.0042} & \multicolumn{1}{c|}{\textbf{0.2837 $\pm$ 0.0043}} & \textbf{0.2414 $\pm$ 0.0018}            & \textbf{0.1981 $\pm$ 0.0014} \\ \hline
\multirow{4}{*}{GraphSAGE} & Vanilla                  & 0.7254 $\pm$ 0.0021          & \multicolumn{1}{c|}{0.5293 $\pm$ 0.0028}          & 0.4503 $\pm$ 0.0014          & \multicolumn{1}{r|}{0.2492 $\pm$ 0.0030}          & \multicolumn{1}{r}{0.2412 $\pm$ 0.0025} & 0.1869 $\pm$ 0.0102          \\
                           & Concat                   & 0.7279 $\pm$ 0.0014          & \multicolumn{1}{c|}{0.5346 $\pm$ 0.0020}          & 0.4694 $\pm$ 0.0045          & \multicolumn{1}{c|}{0.2885 $\pm$ 0.0038}          & 0.2559 $\pm$ 0.0035                     & 0.1986 $\pm$ 0.0002          \\
                           & Addition                 & 0.7284 $\pm$ 0.0005          & \multicolumn{1}{c|}{0.5333 $\pm$ 0.0021}          & 0.4709 $\pm$ 0.0021          & \multicolumn{1}{c|}{0.2870 $\pm$ 0.0061}          & 0.2588 $\pm$ 0.0018                     & 0.2031 $\pm$ 0.0064          \\
                           & LEGNN                    & \textbf{0.7316 $\pm$ 0.0019} & \multicolumn{1}{c|}{\textbf{0.5354 $\pm$ 0.0011}} & \textbf{0.5019 $\pm$ 0.0048} & \multicolumn{1}{c|}{\textbf{0.3098 $\pm$ 0.0042}} & \textbf{0.2857 $\pm$ 0.0031}            & \textbf{0.2306 $\pm$ 0.0009} \\ \hline
\multirow{4}{*}{GAT}       & Vanilla                  & 0.7287 $\pm$ 0.0008          & \multicolumn{1}{c|}{0.5274 $\pm$ 0.0027}          & 0.4898 $\pm$ 0.0047          & \multicolumn{1}{c|}{0.2938 $\pm$ 0.0030}          & 0.2764 $\pm$ 0.0030                     & 0.2165 $\pm$ 0.0013          \\
                           & Concat                   & 0.7313 $\pm$ 0.0019          & \multicolumn{1}{c|}{0.5354 $\pm$ 0.0017}          & 0.5062 $\pm$ 0.0047          & \multicolumn{1}{c|}{0.3235 $\pm$ 0.0030}          & 0.2824 $\pm$ 0.0046                     & 0.2314 $\pm$ 0.0023          \\
                           & Addition                 & 0.7318 $\pm$ 0.0024          & \multicolumn{1}{c|}{0.5372 $\pm$ 0.0012}          & 0.5093 $\pm$ 0.0033          & \multicolumn{1}{c|}{0.3257 $\pm$ 0.0016}          & 0.2899 $\pm$ 0.0035                     & 0.2447 $\pm$ 0.0021          \\
                           & LEGNN                    & \textbf{0.7337 $\pm$ 0.0007} & \multicolumn{1}{c|}{\textbf{0.5397 $\pm$ 0.0009}} & \textbf{0.5276 $\pm$ 0.0014} & \multicolumn{1}{c|}{\textbf{0.3302 $\pm$ 0.0032}} & \textbf{0.3040 $\pm$ 0.0015}            & \textbf{0.2551 $\pm$ 0.0011} \\ \hline
\end{tabular}
}
\end{table*}

From the results, we could conclude that: 1) compared with vanilla GNNs, leveraging the label information could effectively improve the performance, indicating the necessity of integrating labels into GNNs; 2) LEGNN consistently outperforms the existing SOTAs on all the datasets, demonstrating the superiority of LEGNN in utilizing label information; 3) the improvements of LEGNN on datasets with more classes and lower homophily (i.e., ogbn-mag and oag-venue) are more obvious. On the one hand, more classes indicate that labels carry more complicated information. LEGNN explicitly learns the label semantics, and thus performs better. On the other hand, LEGNN enhances the message passing among intra-class nodes even when the graph homophily is low and leads to superior performance. We will verify this assumption in \secref{section-5-synthetic-datasets}.

\subsection{Comparison with SOTAs of Sophisticated Designs}
We employ GAT as the backbone for LEGNN and compare it with state-of-the-arts of sophisticated designs. Due to the different architectures of SOTAs, we strictly report the official results from \citet{DBLP:conf/nips/HuFZDRLCL20} and \citet{yu2022heterogeneous} to make fair comparisons, where the accuracy is adopted as the evaluation metric. The results are shown in \tabref{tab:performance}.
\begin{table}[!htbp]
\centering
\caption{Comparisons with different methods.}
\label{tab:performance}
\begin{tabular}{c|cc}
\hline
Datasets                    & Methods        & Accuracy $\uparrow$       \\ \hline
\multirow{9}{*}{ogbn-arxiv} 
                            & DeeperGCN \cite{li2020deepergcn}    & 0.7192 $\pm$ 0.0016 \\
                            & GaAN \cite{DBLP:conf/uai/ZhangSXMKY18}        & 0.7197 $\pm$ 0.0024 \\
                            & DAGNN \cite{DBLP:conf/kdd/LiuGJ20}       & 0.7209 $\pm$ 0.0025 \\
                            & JKNet \cite{xu2018representation}       & 0.7219 $\pm$ 0.0021 \\
                            & GCNII \cite{chen2020simple}       & 0.7274 $\pm$ 0.0016 \\
                            & UniMP \cite{DBLP:conf/ijcai/ShiHFZWS21}       & 0.7311 $\pm$ 0.0020 \\
                            & MLP + C\&S \cite{DBLP:conf/iclr/HuangHSLB21}  & 0.7312 $\pm$ 0.0012 \\ 
                            & LEGNN   & \textbf{0.7337 $\pm$ 0.0007} \\ 
                            & LEGNN + AS-Train  & \textbf{0.7371 $\pm$ 0.0011}   \\ 
                            \hline
\multirow{9}{*}{ogbn-mag}   
                            & MetaPath2vec \cite{DBLP:conf/kdd/DongCS17} & 0.3544 $\pm$ 0.0036 \\
                            & SIGN \cite{DBLP:journals/corr/abs-2004-11198}        & 0.4046 $\pm$ 0.0012 \\
                            & RGCN \cite{DBLP:conf/esws/SchlichtkrullKB18}        & 0.4678 $\pm$ 0.0067 \\
                            & HGT \cite{DBLP:conf/www/HuDWS20}         & 0.4927 $\pm$ 0.0061 \\
                            & R-GSN \cite{DBLP:journals/corr/abs-2103-07877}       & 0.5032 $\pm$ 0.0037 \\
                            & HGConv \cite{DBLP:journals/corr/abs-2012-14722}       & 0.5045 $\pm$ 0.0017 \\
                            & LEGNN   & \textbf{0.5276 $\pm$ 0.0014} \\ 
                            & LEGNN + AS-Train   & \textbf{0.5378 $\pm$ 0.0016} \\ 
                            \hline
\multirow{7}{*}{oag-venue}  & RSHN \cite{DBLP:conf/icdm/ZhuZPZW19}        & 0.2159 $\pm$ 0.0023          \\
                            & RGCN \cite{DBLP:conf/esws/SchlichtkrullKB18}        & 0.2397 $\pm$ 0.0013          \\
                            & HGT \cite{DBLP:conf/www/HuDWS20}          & 0.2447 $\pm$ 0.0019          \\
                            & HetSANN \cite{DBLP:conf/aaai/HongGLYLY20}     & 0.2581 $\pm$ 0.0021         \\
                            & R-HGNN \cite{yu2022heterogeneous}      & 0.2887 $\pm$ 0.0012          \\ 
                            & LEGNN   & \textbf{0.3040 $\pm$ 0.0015} \\ 
                            & LEGNN + AS-Train   & \textbf{0.3086 $\pm$ 0.0018} \\ 
                            \hline
\end{tabular}
\end{table}

From \tabref{tab:performance}, we could observe that although the SOTAs utilize more sophisticated designs than GAT, our method still obtains superior performance on all the datasets with GAT as the backbone. This phenomenon reveals the effectiveness of our approach in integrating rich label information into GNNs. Moreover, when equipped with AS-Train, the performance is improved further, which proves the advantage of AS-Train due to its ability in providing more reliable pseudo labels and distinguishing the importance of each pseudo-labeled node.

\subsection{Experiments on Synthetic Datasets}
\label{section-5-synthetic-datasets}
\begin{figure}[!htbp]
    \centering
    \includegraphics[scale=0.385]{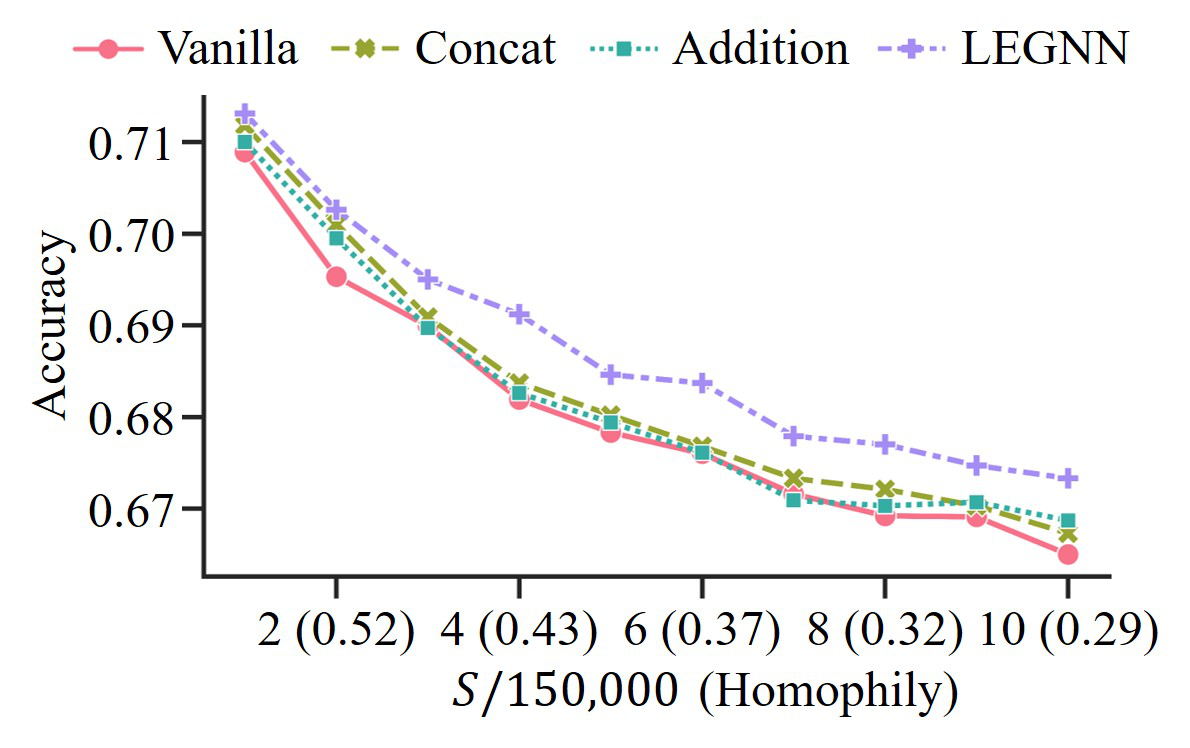}
\caption{Performance of different methods on syn-arxiv.}
\label{fig:synthetic_result}
\end{figure}

We also generate the synthetic dataset called \textbf{syn-arxiv} based on ogbn-arxiv to validate the effectiveness of our approach in smoothing the representations of intra-class nodes. We modify the graph structure of ogbn-arxiv by adding cross-label edges to connect nodes with different labels. In particular, we first randomly sample a node $v_i$ from the node set $\mathcal{V}$ and obtain the label $y_i$ of node $v_i$. Then, we randomly sample another node $v_j$ from $\mathcal{V}$ whose label $y_j$ is different $y_i$. Next, we add an edge that connects $v_i$ and $v_j$ to the edge set $\mathcal{E}$. Finally, we add a total number of $S$ edges to $\mathcal{E}$ and obtain the synthetic dataset syn-arxiv. In \appendixref{section-appendix-synthetic-statistic}, \tabref{tab:synthetic_homophily} shows the values of $S$ and graph homophily on synthetic datasets. We then compare LEGNN with the existing methods on syn-arxiv and report the performance of accuracy in \figref{fig:synthetic_result}.

From \figref{fig:synthetic_result}, we find that LEGNN could consistently outperform the existing methods across various settings of $S$. Moreover, the improvements of our approach over the compared methods become \textit{more obvious with the increase of $S$}. This is because when $S$ increases, the graph homophily would decrease. Our LEGNN establishes the edges between nodes and labels and thus enhances the message passing among intra-class nodes. This makes LEGNN benefit more than the baselines when the graph homophily is lower.

\subsection{In-Depth Analysis on LEGNN}
We then investigate why LEGNN could improve the performance of node classification. We first define the Label Difference (LD) and Graph Difference (GD). LD is defined as the difference of intra-class nodes with their readout. Specifically, the label difference of class $c$ is calculated by,
\begin{equation}
    LD_c = \frac{1}{|\mathcal{V}_c|} \sum_{v \in \mathcal{V}_c}\|\bm{z}_v - \bm{z}_c\|_2,
\end{equation}
where $\mathcal{V}_c$ denotes the set of nodes belonging to class $c$. $\bm{z}_c$ is the readout of nodes with class $c$, and we use the average pooling to derive it as follows,
\begin{equation}
    \bm{z}_c = \frac{1}{|\mathcal{V}_c|}\sum_{v \in \mathcal{V}_c} \bm{z}_v.
\end{equation}
GD is defined as the average of LD of the all label classes,
\begin{equation}
    GD = \frac{1}{C} \sum_{c=1}^{C} LD_c,
\end{equation}
where $C$ is the number of label classes. Notably, lower GD indicates that intra-class node representations are smoother. We calculate GD of different methods on ogbn-arxiv and show the results in \tabref{tab:cluster_graph_smoothness}.

\begin{table}[!htbp]
\centering
\caption{Results of GD and node clustering on ogbn-arxiv.}
\label{tab:cluster_graph_smoothness}
\begin{tabular}{c|ccc}
\hline
\multirow{2}{*}{Methods} & \multicolumn{3}{c}{Metrics}                          \\ \cline{2-4} 
                        & GD $\downarrow$          & NMI $\uparrow$             & ARI $\uparrow$ \\ \hline
Vanilla                  & 12.4364      & 0.7058          & 0.6631                    \\
Concat                  & 11.6597      & 0.7092          & 0.6626                  \\
Addition                & 11.5814      & 0.7168          & 0.6752                  \\
LEGNN             & \textbf{8.8851}    & \textbf{0.7324} & \textbf{0.7031}   \\ \hline
\end{tabular}
\end{table}

From \tabref{tab:cluster_graph_smoothness}, we could observe that GD is significantly reduced by LEGNN compared with other methods. This reveals the superiority of LEGNN over the existing methods lies in the smoothing of intra-class node representations, which is beneficial for classifying nodes.

\begin{figure}[!htbp]
    \centering
    \subfigure[Vanilla.]{
        \includegraphics[width=.465\columnwidth]{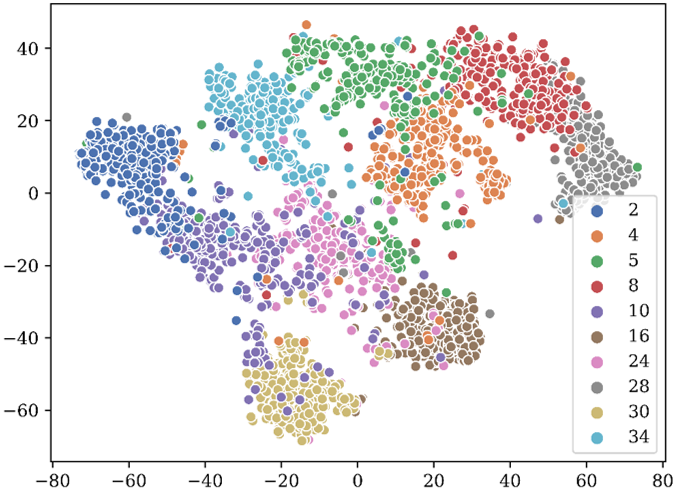}
        \label{fig:visualization_origin}
    }
    \subfigure[Concat.]{
        \includegraphics[width=.465\columnwidth]{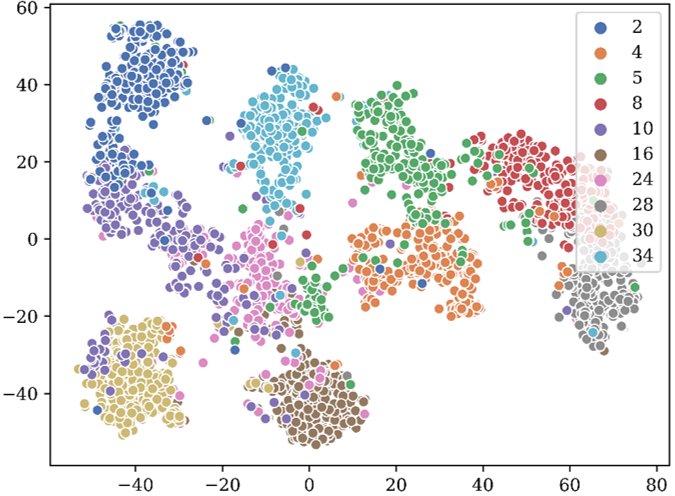}
        \label{fig:visualization_concat}
    }
    \subfigure[Addition.]{
        \includegraphics[width=.47\columnwidth]{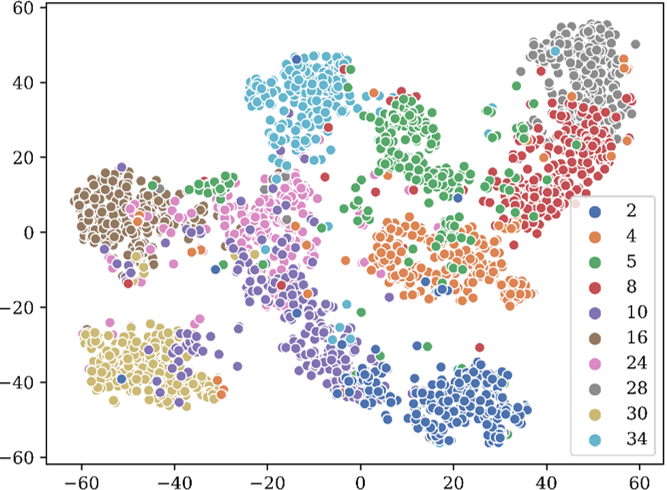}
        \label{fig:visualization_addition}
    }
    \subfigure[LEGNN.]{
        \includegraphics[width=.47\columnwidth]{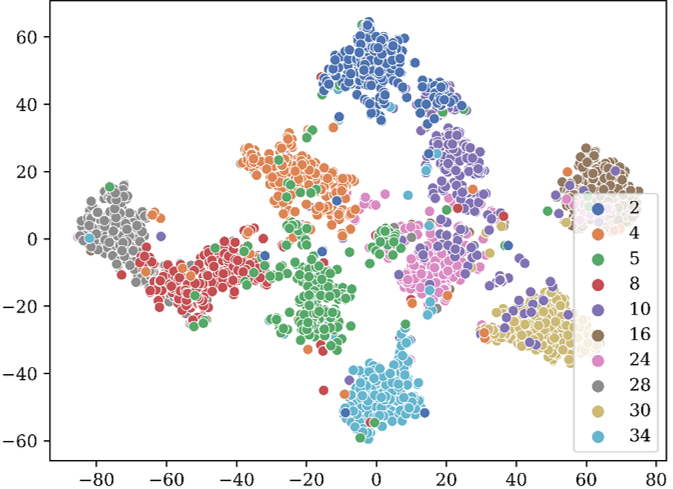}
        \label{fig:visualization_labelasnode}
    }
    \caption{Node visualization on ogbn-arxiv.}
    \label{fig:visualization}
\end{figure}

We conduct the node clustering and node visualization tasks to further validate this advantage of LEGNN. We first choose the top 10 classes of papers and then randomly select 500 papers from each class. Therefore, we obtain 5,000 papers in total. Then, we feed the selected 5,000 papers into k-means and t-SNE \cite{maaten2008visualizing} (which projects node representations into a 2-dimensional space) to get the node clustering and node visualization results. NMI and ARI are adopted as the evaluation metrics for node clustering. 
Experimental results are shown in \tabref{tab:cluster_graph_smoothness} and \figref{fig:visualization}.

In \tabref{tab:cluster_graph_smoothness}, LEGNN obtains the best performance on node clustering due to its effectiveness in enhancing the intra-class node representation smoothness. \figref{fig:visualization} shows that LEGNN performs better than baselines on node visualization. Compared with baselines, LEGNN gathers nodes in the same class more closely and provides more obvious boundaries between nodes in different classes.

We also show the gain of using label feature matrix $\bm{E}$ for learning label semantics. We replace each label's feature with the average of the features of training nodes belonging to the corresponding label. This replacement makes the performance of LEGNN drop from 0.7329 to 0.7315, from 0.7316 to 0.7292, from 0.7337 to 0.7325 with GCN, GraphSAGE and GAT as the backbone, respectively. This validates the benefit of leveraging label feature matrix $\bm{E}$ in LEGNN.

\subsection{Necessity of Training Node Selection}
\label{section-5-training-node-selection}
We conduct experiments on ogbn-arxiv with GAT as the backbone to show the necessity of using the Training Node Selection (TNS). We remove TNS by establishing the connections between all the labeled nodes with their labels and training the model to predict all the labeled nodes. Results are reported in \tabref{tab:training_node_selection}.

\begin{table}[!htbp]
\centering
\caption{Effects of the training node selection technique.}
\label{tab:training_node_selection}
\resizebox{1.02\columnwidth}{!}{
\setlength{\tabcolsep}{1.0mm}{
\begin{tabular}{c|ccc|ccc}
\hline
\multirow{2}{*}{Methods} & \multicolumn{3}{c|}{Accuracy $\uparrow$}  & \multicolumn{3}{c}{Macro-F1 $\uparrow$}  \\ \cline{2-7} 
                         & Train    & Validate & Test     & Train   & Validate & Test     \\ \hline
Vanilla                  & 0.8479   & 0.7427   & 0.7287   & 0.7542  & 0.5507   & 0.5274   \\
\multirow{2}{*}{w/o TNS} & \textbf{0.9607}   & 0.6824   & 0.6752   & \textbf{0.8821}  & 0.4040   & 0.3839   \\
                         & +13.30\% & -8.12\% & -7.34\% & +16.96\% & -26.64\% & -27.21\% \\
\multirow{2}{*}{w TNS}   & 0.8466   & \textbf{0.7484}   & \textbf{0.7337}   & 0.7527  & \textbf{0.5637}   & \textbf{0.5397}   \\
                         & -0.15\%  & +0.77\%  & +0.69\%  & -0.20\% & +2.36\%  & +2.33\%  \\ \hline
\end{tabular}
}
}
\end{table}

From \tabref{tab:training_node_selection}, we could observe that: 1) removing TNS makes the model get much higher metrics on the training set but leads to poor performance on the validation or test metrics, which is caused by the label leakage issue; 2) using TNS prevents the model from overfitting the training data and effectively improves the model generalization ability.

\subsection{Ablation Study on AS-Train}
\begin{figure}[!htbp]
    \centering
    \includegraphics[width=\columnwidth]{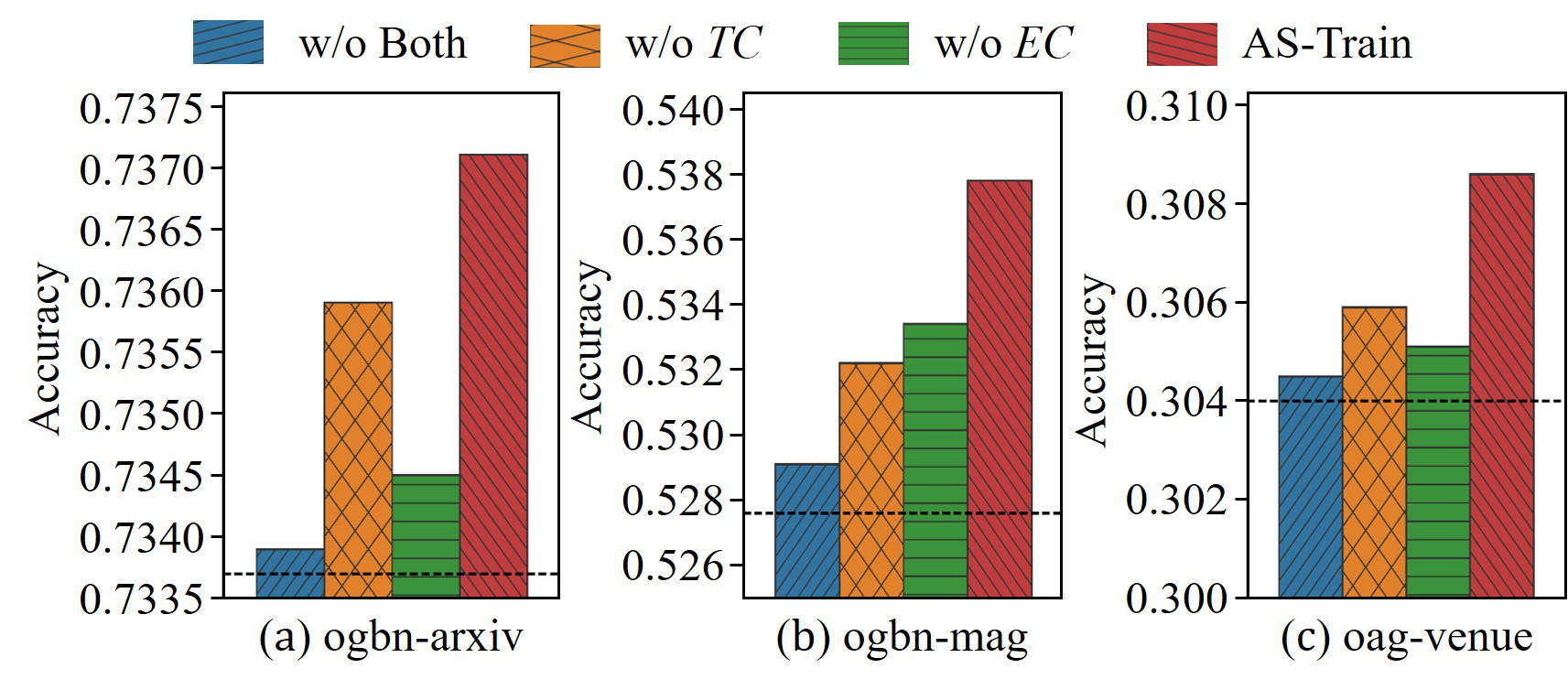}
\caption{Effects of the $TC$ and $EC$ in AS-Train.}
\label{fig:ablation_study}
\end{figure}

We also investigate the effects of the training confidence $TC$ and evaluating confidence $EC$ in AS-Train. Specifically, we use w/o $TC$, w/o $EC$, and w/o Both to denote the remove of training confidence, evaluating confidence, and both of them, respectively. It is worth noticing that the w/o Both variant is equal to the previous methods for self-training on graphs (e.g., \cite{zhou2019effective,sun2021scalable}) that only use a pre-defined threshold to select pseudo-labeled nodes. \figref{fig:ablation_study} shows the performance of different variants on the three datasets, where the dotted black line denotes the performance of LEGNN without using the self-training strategy. 

From \figref{fig:ablation_study}, we observe that both $TC$ and $EC$ contribute to the improvements in performance and removing any of them would lead to worse results. Concretely, $TC$ improves the reliability of pseudo labels and $EC$ distinguishes the importance of each pseudo-labeled node. The w/o Both variant achieves the worst performance. Compared with LEGNN without using the self-training strategy, w/o Both shows minor improvements on the three datasets. This phenomenon indicates that trivially using the threshold-based technique is insufficient \cite{zhou2019effective,sun2021scalable} and it is necessary to make careful designs for optimally leveraging the benefits of the self-training strategy.

\subsection{Computational Cost Comparison}
\begin{figure}[!htbp]
    \centering
    \includegraphics[width=1.02\columnwidth]{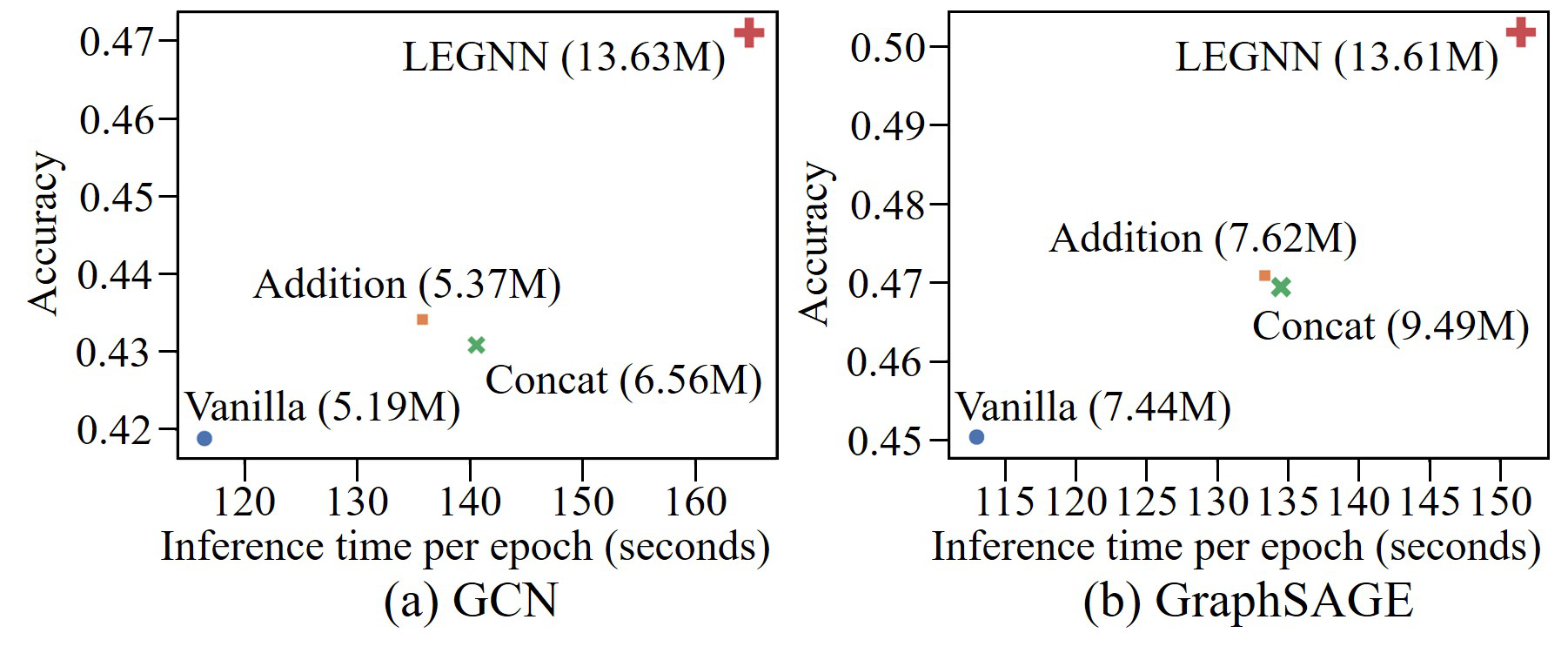}
    \caption{Comparisons of the parameter size and inference time of different methods on ogbn-mag.}
    \label{fig:model_computational_cost}
\end{figure}

We further compare the computational cost of our method with baselines (i.e., Vanilla GNNs, Concat and Addition). We report the inference time to eliminate the effects of different training strategies of the methods. Due to space limitations, we only show the results of GCN and GraphSAGE backbones on the largest ogbn-mag dataset in \figref{fig:model_computational_cost}, and similar observations can be found on the GAT backbone.

From \figref{fig:model_computational_cost}, we find that compared with the baselines, LEGNN achieves 9.20\% improvements in accuracy with 1.23$\times$ increase in inference time and 2.05$\times$ increase in parameter capacity on average. Overall, although our method inevitably costs more time in learning label representations and introduces additional parameters for labels, it still obtains a good trade-off between effectiveness and efficiency.

\subsection{Parameter Sensitivity}
\begin{figure}[!htbp]
    \centering
    \subfigure[Effects of $\alpha$.]{
        \includegraphics[width=0.474\columnwidth]{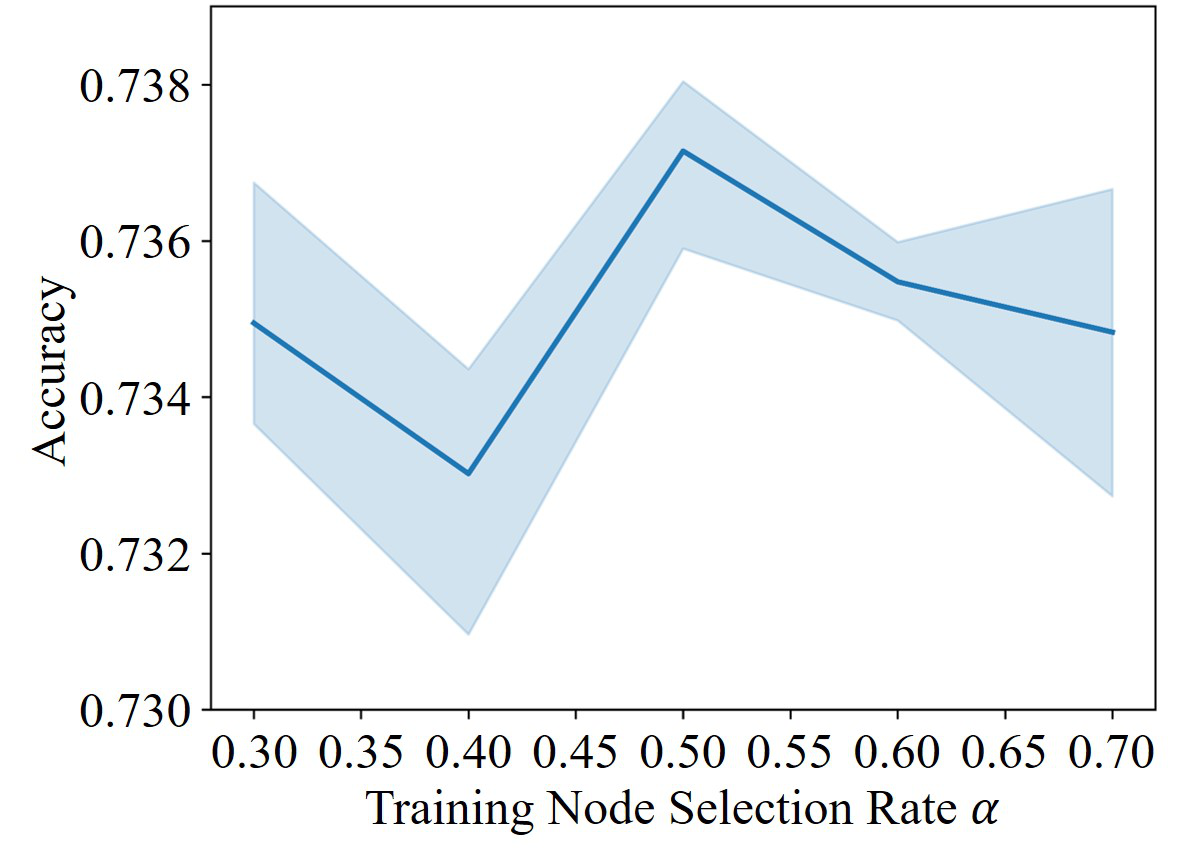}
        \label{fig:parameter_mask_rate}
    }
    \subfigure[Effects of $\delta$.]{
        \includegraphics[width=0.474\columnwidth]{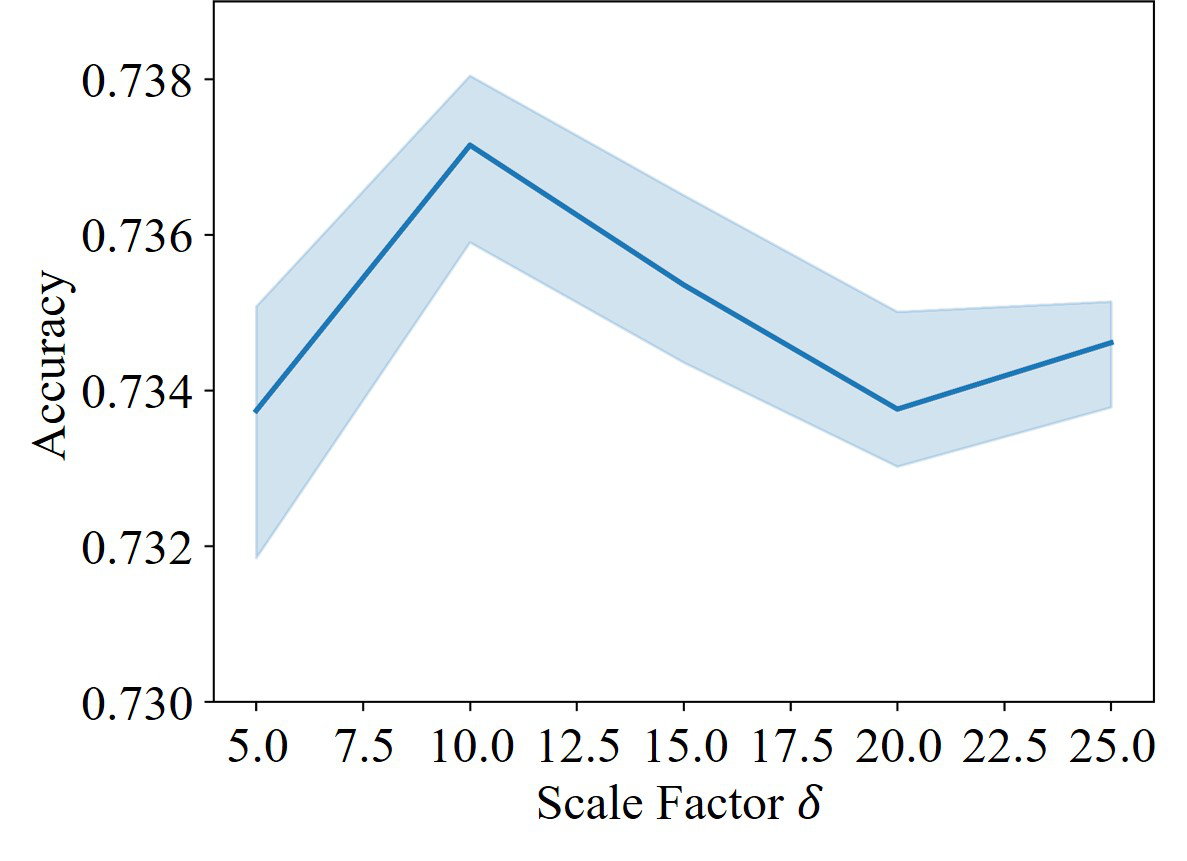}
        \label{fig:parameter_temperature}
    }
    \subfigure[Effects of $\lambda$.]{
        \includegraphics[width=0.474\columnwidth]{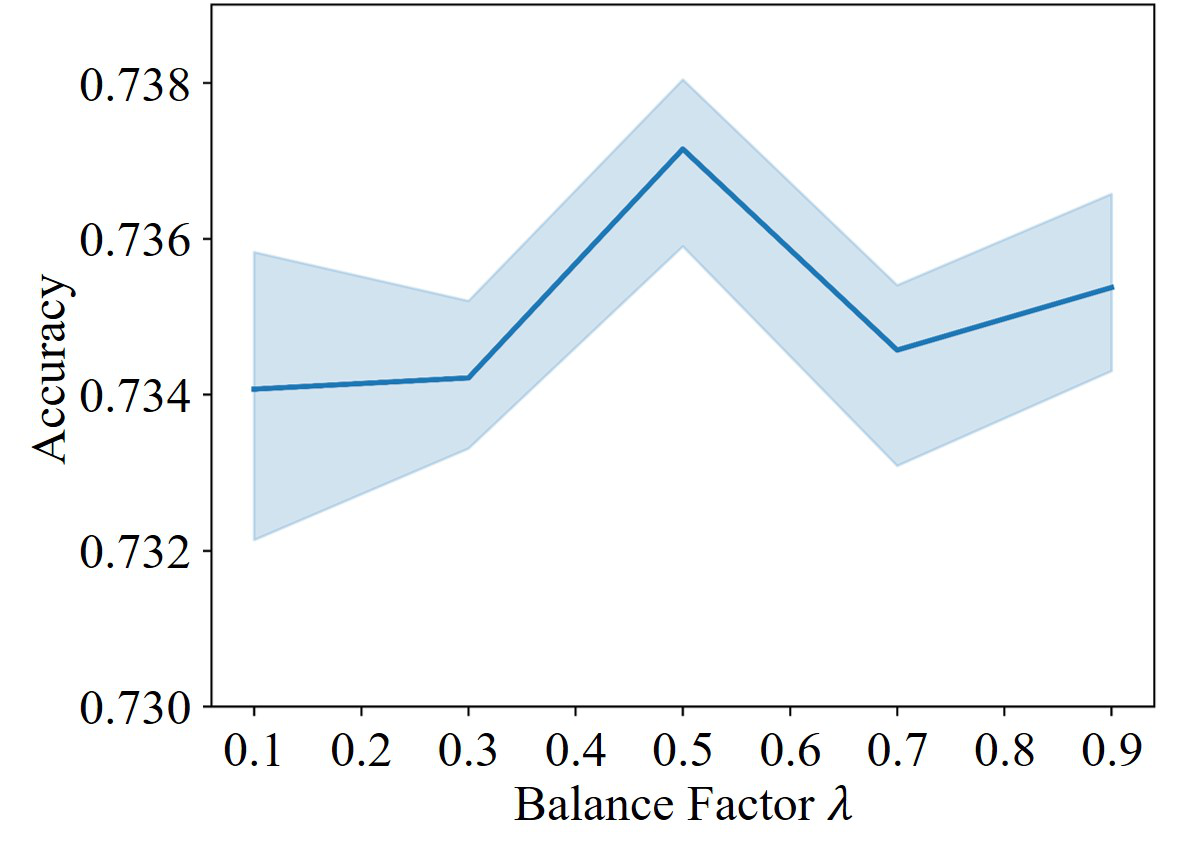}
        \label{fig:parameter_balance_factor}
    }
    \subfigure[Effects of $D$.]{
        \includegraphics[width=0.474\columnwidth]{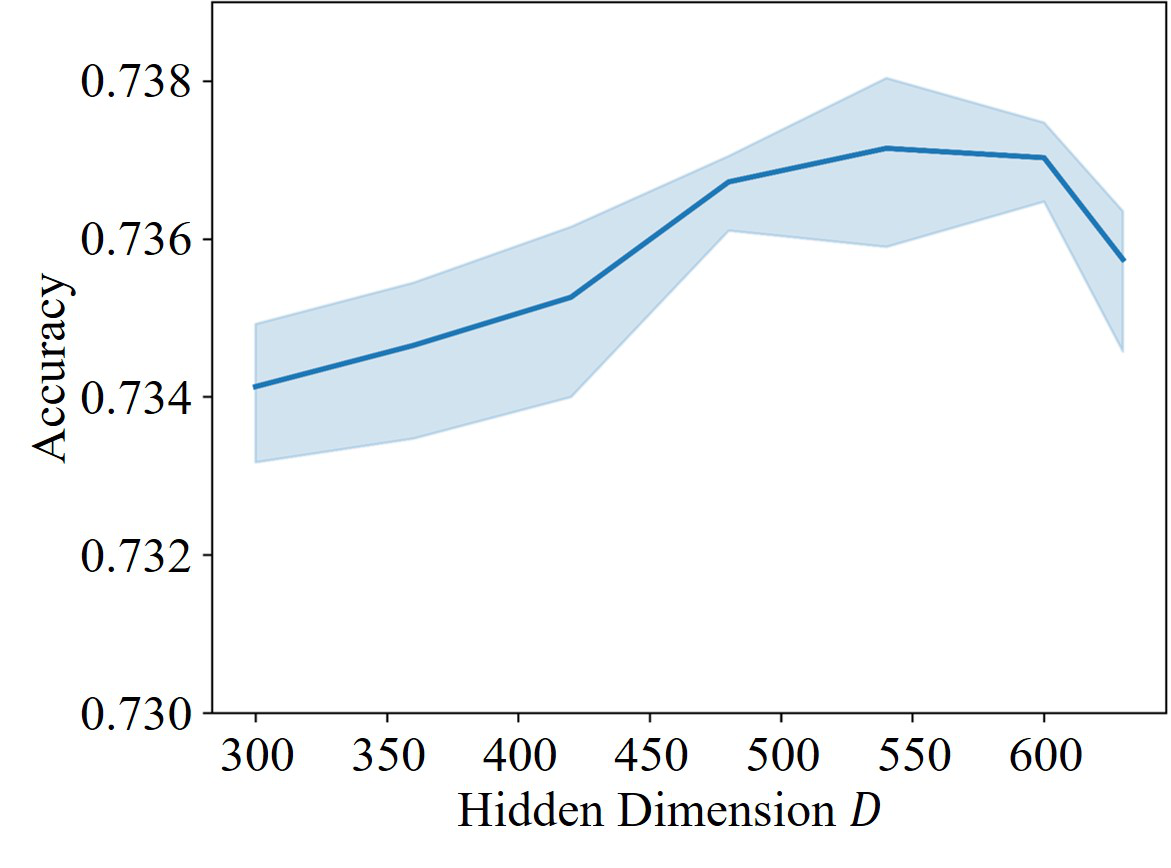}
        \label{fig:parameter_hidden_dimension}
    }
    \caption{Hyperparameter analysis on $\alpha$, $\delta$, $\lambda$ and $D$.}
    \label{fig:parameter_sensitivity}
\end{figure}

We also investigate how do the hyperparameters affect the model performance, including the training node selection rate $\alpha$, the scale factor $\delta$, the balance factor $\lambda$ of pseudo labels and the hidden dimension $D$. We vary the settings of the hyperparameters and show the results in \figref{fig:parameter_sensitivity}.

\figref{fig:parameter_mask_rate} indicates that it is essential to appropriately determine the proportion of labeled nodes for establishing edges with labels and making predictions. \figref{fig:parameter_temperature} proves that the better $\delta$ approximates the accuracy convergence curve, the higher performance the model would achieve (i.e., $\delta=10$ optimally approximates the accuracy convergence curve in \figref{fig:different_temperature_log}). \figref{fig:parameter_balance_factor} shows the necessity to control the importance of pseudo-labeled nodes by suitable values of $\lambda$ for a good balance. From \figref{fig:parameter_hidden_dimension}, we find that the performance of LEGNN grows with the increment of the hidden dimension $D$ and obtains the best performance when $D$ is set to 540. However, the performance drops when $D$ gets larger further, which may be caused by the over-fitting problem because of too many parameters.

\section{Conclusion}
\label{section-6}

In this paper, we proposed a label-enhanced learning framework to comprehensively integrate rich label information into GNNs for improving semi-supervised node classification. Our approach created a virtual center for each label and jointly learned representations of both nodes and labels with the heterogeneous message passing mechanism. Our method could effectively smooth the smoothness of intra-class node representations and explicitly encode label semantics in the learning process of GNNs. A training node selection technique is further introduced to tackle the label leakage issue and improve the model generalization ability. We also designed an adaptive self-training strategy to provide more reliable pseudo labels and distinguish the importance of each pseudo-labeled node. We conducted extensive experiments on real-world and synthetic datasets, and the results demonstrated the superiority of our approach over the existing methods.

\ifCLASSOPTIONcompsoc
  \section*{Acknowledgments}
\else
  \section*{Acknowledgment}
\fi
This work was supported in part by the National Key R$\&$D Program of China [grant number 2021YFB2104802], and the National Natural Science Foundation of China [grant numbers 62272023, 71901011 and 51991395].

\ifCLASSOPTIONcaptionsoff
  \newpage
\fi



\bibliographystyle{IEEEtran}
\bibliography{reference}

\begin{IEEEbiography}[{\includegraphics[width=1in,height=1.25in,clip,keepaspectratio]{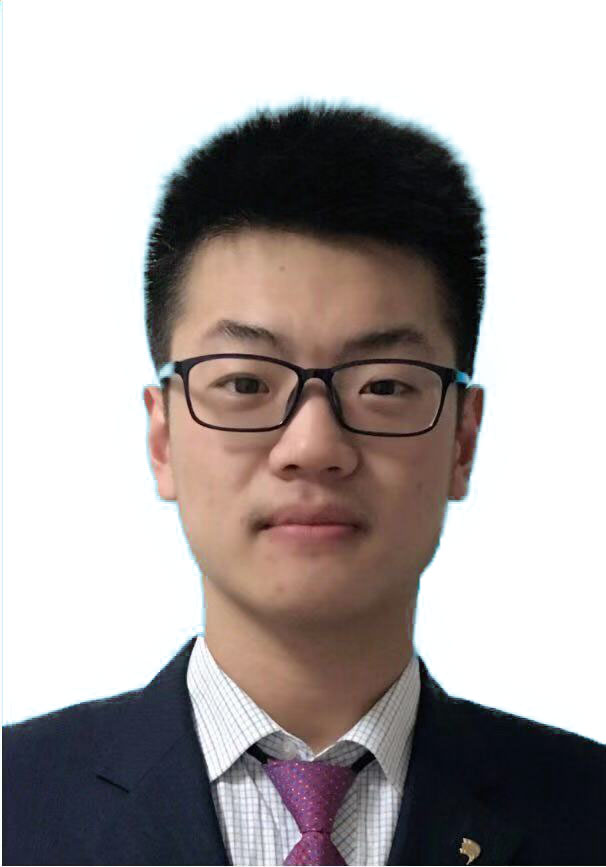}}]
{Le Yu} received the B.S. degree in Computer Science and Engineering from Beihang University, Beijing, China, in 2019. He is currently a third-year computer science Ph.D. student in the School of Computer Science and Engineering at Beihang University. His research interests include temporal data mining, machine learning and graph neural networks.
\end{IEEEbiography}

\begin{IEEEbiography}[{\includegraphics[width=1in,height=1.25in,clip,keepaspectratio]{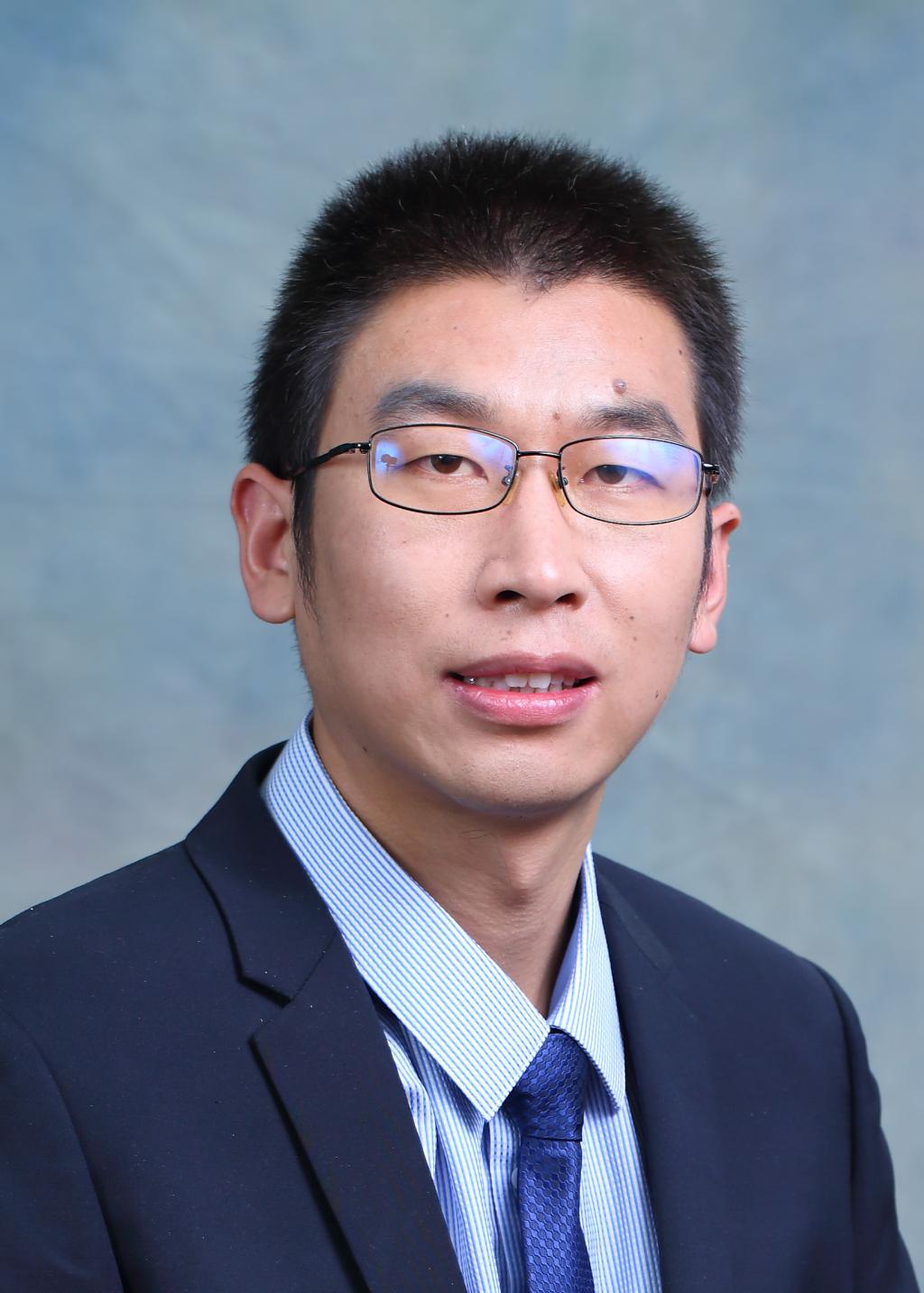}}]
{Leilei Sun} is currently an associate professor in School of Computer Science, Beihang University, Beijing, China. He was a postdoctoral research fellow from 2017 to 2019 in School of Economics and Management, Tsinghua University. He received his Ph.D. degree from Institute of Systems Engineering, Dalian University of Technology, in 2017. His research interests include machine learning and data mining. 
\end{IEEEbiography}

\begin{IEEEbiography}[{\includegraphics[width=1in,height=1.25in,clip,keepaspectratio]{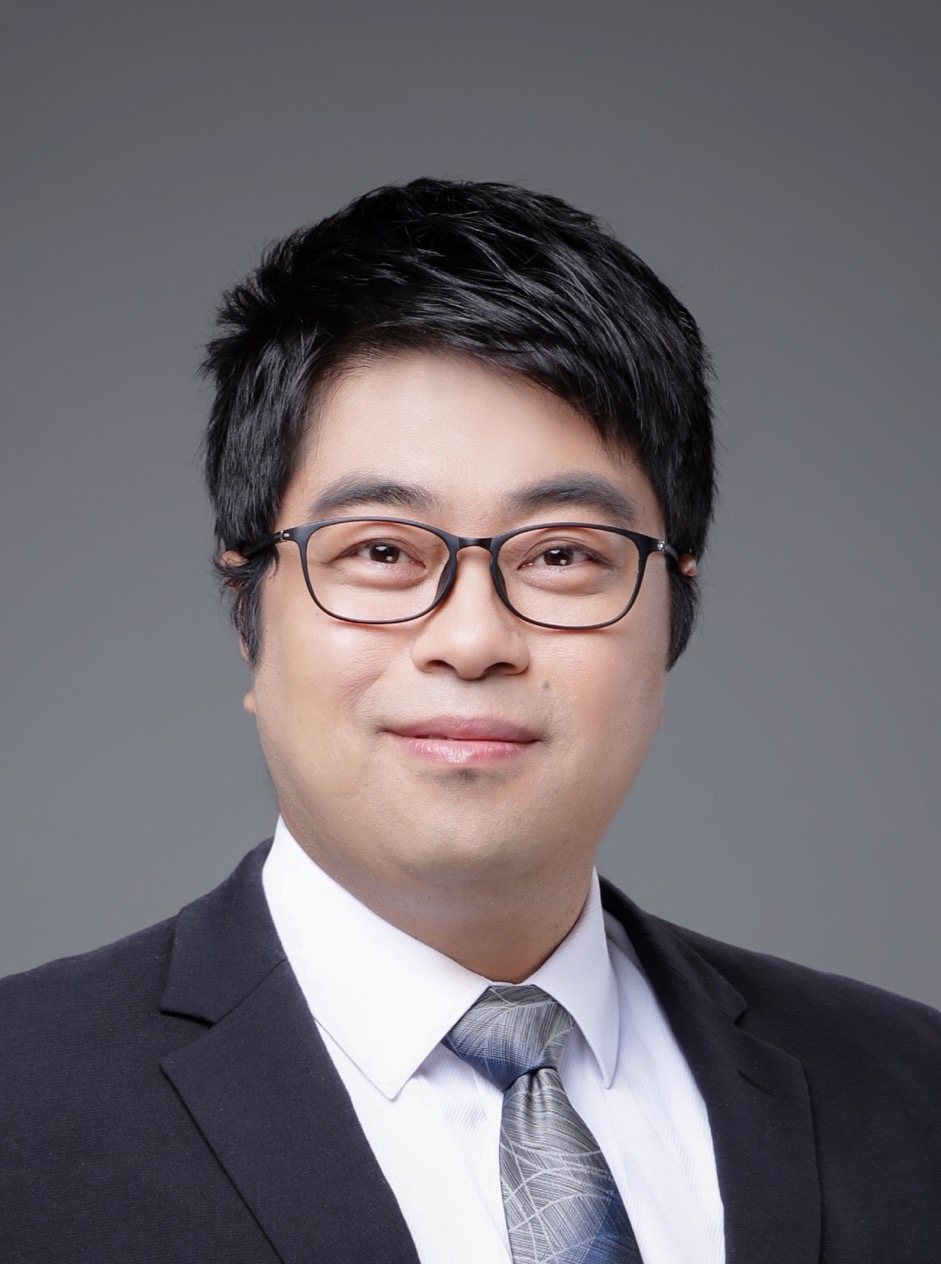}}]
{Bowen Du} received the Ph.D. degree in Computer Science and Engineering from Beihang University, Beijing, China, in 2013. He is currently a Professor with the State Key Laboratory of Software Development Environment, Beihang University. His research interests include smart city technology, multi-source data fusion, and traffic data mining.
\end{IEEEbiography}

\begin{IEEEbiography}[{\includegraphics[width=1in,height=1.25in,clip,keepaspectratio]{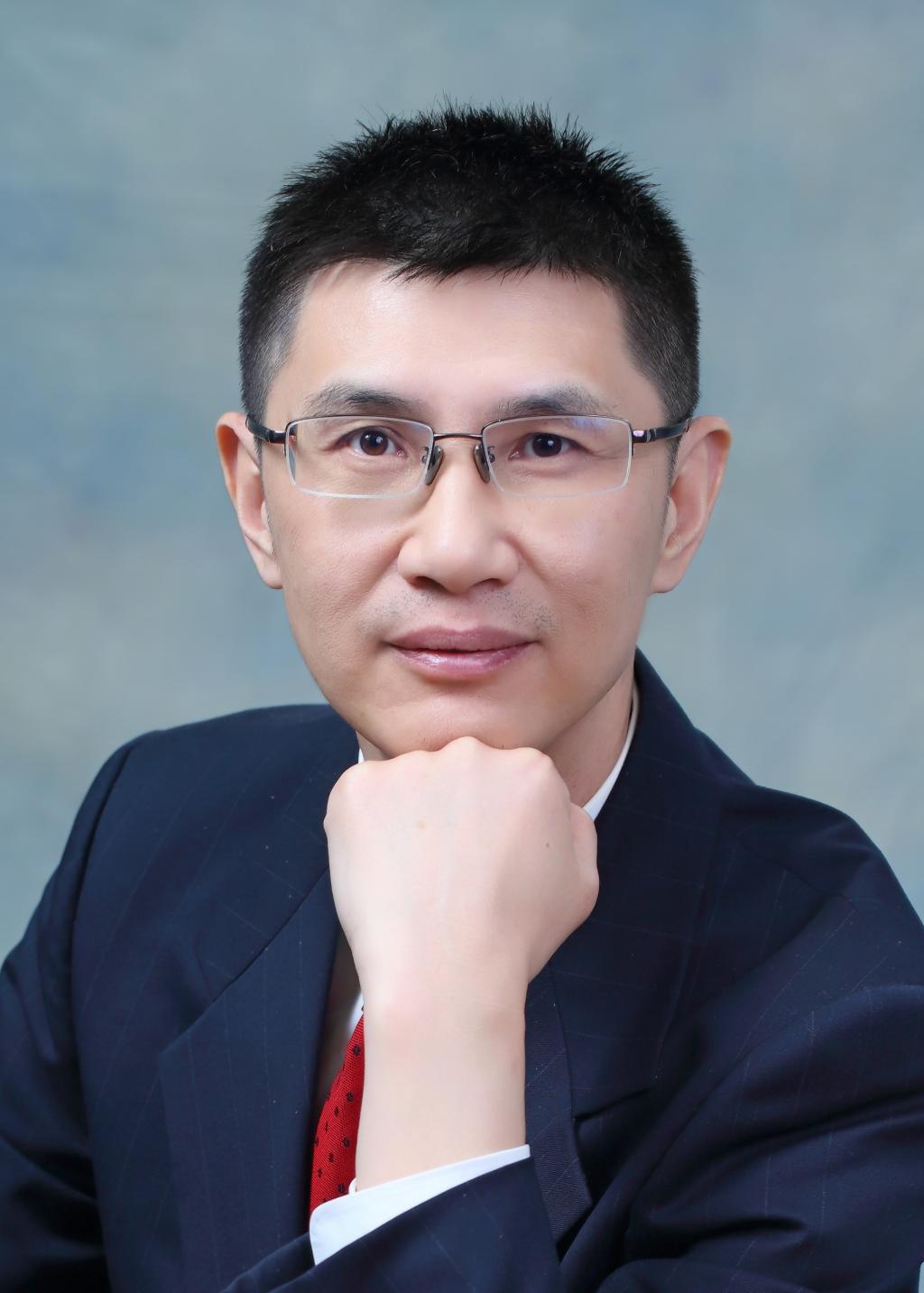}}]
{Tongyu Zhu} received the B.S. degree from Tsinghua University in 1992, and the M.S. degree from Beihang University in 1999. He is currently an associate professor with the State Key Laboratory of Software Development Environment in School of Computer Science, Beihang University, Beijing, China. His research interests include intelligent traffic information processing and network application.
\end{IEEEbiography}

\begin{IEEEbiography}[{\includegraphics[width=1in,height=1.25in,clip,keepaspectratio]{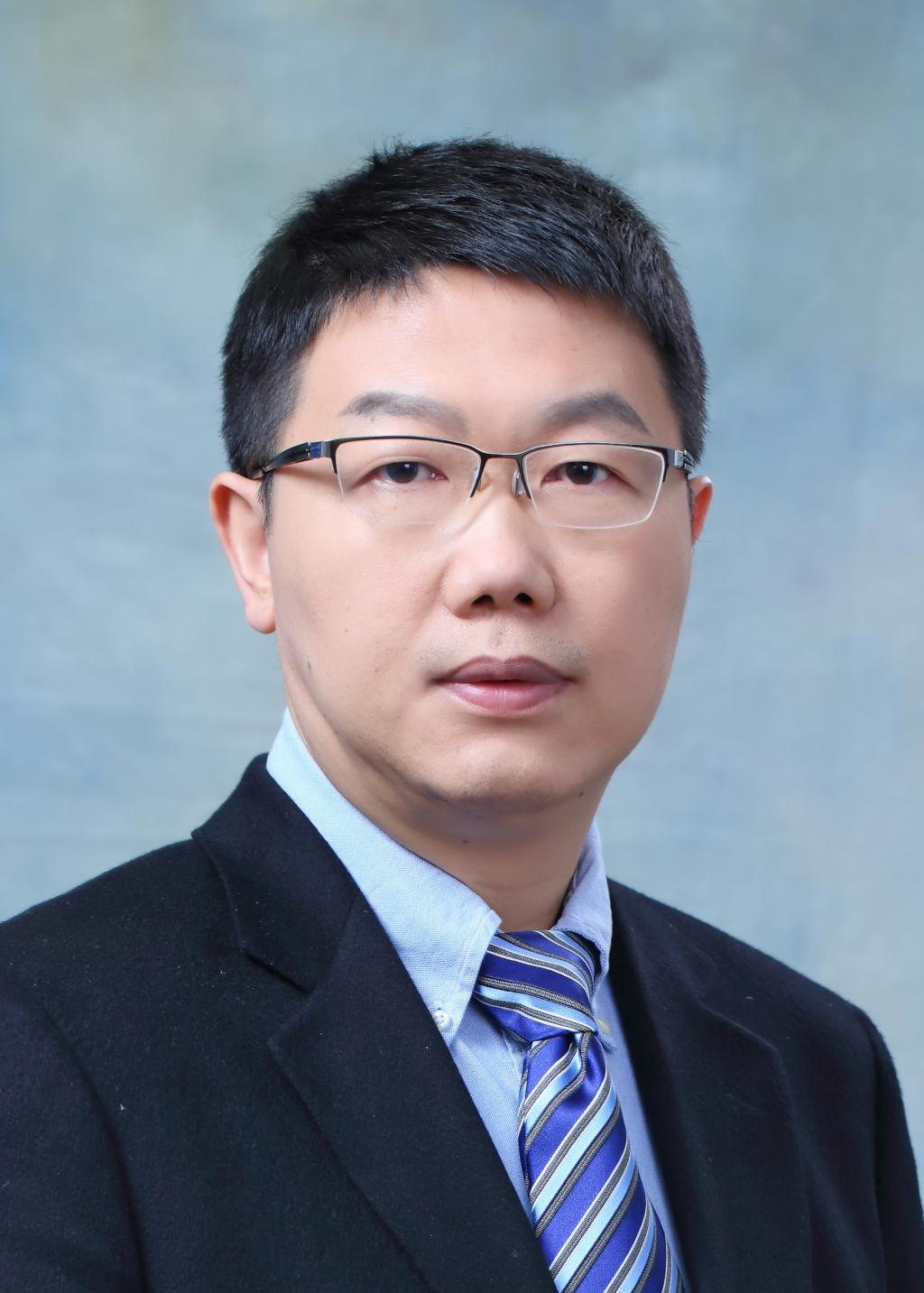}}]
{Weifeng Lv} received the B.S. degree in Computer Science and Engineering from Shandong University, Jinan, China, and the Ph.D. degree in Computer Science and Engineering from Beihang University, Beijing, China, in 1992 and 1998 respectively. Currently, he is a Professor with the State Key Laboratory of Software Development Environment, Beihang University, Beijing, China. His research interests include smart city technology and mass data processing.
\end{IEEEbiography}

\clearpage
\appendix
\label{section-appendix}
In the appendix, details of the experiments are introduced.

\subsection*{Settings of Dropout and Learning Rate} \label{section-appendix-dropout-learning_rate}
The settings of dropout and learning rate on all the methods are shown in \tabref{tab:dropout_learning_rate}.

\begin{table}[!ht]
\centering
\caption{Settings of dropout and learning rate of baselines and our approach.}
\label{tab:dropout_learning_rate}
\begin{tabular}{c|c|cccccc}
\hline
\multirow{3}{*}{Backbones} & \multirow{3}{*}{Methods} & \multicolumn{6}{c}{Datasets}                                                                                                \\ \cline{3-8} 
                           &                          & \multicolumn{2}{c|}{ogbn-arxiv}              & \multicolumn{2}{c|}{ogbn-mag}                & \multicolumn{2}{c}{oag-venue} \\ \cline{3-8} 
                           &                          & dropout & \multicolumn{1}{c|}{learning rate} & dropout & \multicolumn{1}{c|}{learning rate} & dropout    & learning rate    \\ \hline
\multirow{4}{*}{GCN}       & Vanilla                  & 0.6     & \multicolumn{1}{c|}{0.001}         & 0.3     & \multicolumn{1}{c|}{0.001}         & 0.2        & 0.001            \\
                           & Concat                   & 0.4     & \multicolumn{1}{c|}{0.001}         & 0.3     & \multicolumn{1}{c|}{0.001}         & 0.2        & 0.001            \\
                           & Addition                 & 0.6     & \multicolumn{1}{c|}{0.001}         & 0.3     & \multicolumn{1}{c|}{0.001}         & 0.2        & 0.001            \\
                           & LEGNN                    & 0.6     & \multicolumn{1}{c|}{0.002}         & 0.3     & \multicolumn{1}{c|}{0.001}         & 0.4        & 0.001            \\ \hline
\multirow{4}{*}{GraphSAGE} & Vanilla                  & 0.6     & \multicolumn{1}{c|}{0.001}         & 0.3     & \multicolumn{1}{c|}{0.001}         & 0.3        & 0.001            \\
                           & Concat                   & 0.5     & \multicolumn{1}{c|}{0.001}         & 0.4     & \multicolumn{1}{c|}{0.001}         & 0.2        & 0.001            \\
                           & Addition                 & 0.6     & \multicolumn{1}{c|}{0.002}         & 0.4     & \multicolumn{1}{c|}{0.001}         & 0.3        & 0.001            \\
                           & LEGNN                    & 0.5     & \multicolumn{1}{c|}{0.002}         & 0.3     & \multicolumn{1}{c|}{0.001}         & 0.4        & 0.001            \\ \hline
\multirow{4}{*}{GAT}       & Vanilla                  & 0.6     & \multicolumn{1}{c|}{0.001}         & 0.4     & \multicolumn{1}{c|}{0.001}         & 0.2        & 0.001            \\
                           & Concat                   & 0.6     & \multicolumn{1}{c|}{0.001}         & 0.4     & \multicolumn{1}{c|}{0.001}         & 0.2        & 0.001            \\
                           & Addition                 & 0.6     & \multicolumn{1}{c|}{0.002}         & 0.3     & \multicolumn{1}{c|}{0.001}         & 0.3        & 0.001            \\
                           & LEGNN                    & 0.6     & \multicolumn{1}{c|}{0.002}         & 0.3     & \multicolumn{1}{c|}{0.001}         & 0.4        & 0.001            \\ \hline
\end{tabular}
\end{table}

\subsection*{Settings of Hyperparameters} \label{section-appendix-hyper-parameters}
\tabref{tab:hyperparameter_settings} shows the hyperparameter settings of our approach.

\begin{table}[!ht]
\centering
\caption{Hyperparameter settings of our approach on different datasets.}
\label{tab:hyperparameter_settings}
\begin{tabular}{c|cccc}
\hline
Datasets   & $\alpha$   & $\delta$  &  $t$  & $\lambda$ \\ \hline
ogbn-arxiv & 0.50   & 10.0  &  0.70 & 0.50 \\
ogbn-mag   & 0.50 & 5.0 &  0.60 &  0.10  \\
oag-venue  & 0.50   & 30.0 &  0.90  & 0.05 \\ \hline
\end{tabular}
\end{table}

\subsection*{Details of Synthetic Datasets} \label{section-appendix-synthetic-statistic}
\tabref{tab:synthetic_homophily} shows the values of $S$ and graph homophily on synthetic datasets.

\begin{table}[!ht]
\centering
\caption{The values of $S$ and graph homophily on synthetic datasets.}
\label{tab:synthetic_homophily}
\begin{tabular}{c|cccccccccc}
\hline
$S$         & 150,000 & 300,000 & 450,000 & 600,000 & 750,000 & 900,000 & 1,050,000 & 1,200,000 & 1,350,000 & 1,500,000 \\ \hline
Homophily & 0.5804  & 0.5211  & 0.4727  & 0.4325  & 0.3987  & 0.3697  & 0.3447    & 0.3229    & 0.3036    & 0.2865    \\ \hline
\end{tabular}
\end{table}

\end{document}